\documentclass[num-refs]{wiley-article}

\usepackage{color}
\usepackage{ulem}

\usepackage{amssymb,amsopn, amsxtra} 
\usepackage{amsmath}
\usepackage{natbib}
\usepackage{graphicx,graphics,subfigure}
\usepackage{siunitx}

\newtheorem{condition}{Condition}

\newcommand{\mr}[1]{\mathrm{#1}}
\newcommand{\mb}[1]{\mathbf{#1}}
\newcommand{\mc}[1]{\mathcal{#1}}

\newcommand\Cov{\mathsf{Cov}}
\newcommand\E{\mathsf{E}}

\newcommand{\tr}[1]{\mr{tr}\left(#1\right)}

\renewcommand\Pr{\mathsf{Pr}}

\newcommand{\refp}[1]{(\ref{#1})}

\newcommand{\per}{\tilde{\mr{f}}}       
\newcommand{\PER}{\tilde{\mb{f}}}
\newcommand{\rad}{\mr{r}}
\newcommand{\vect}[1]{\mr{vec}{#1}}

\newcommand{\boldbeta}{\boldsymbol \beta}
\newcommand{\bolddelta}{\boldsymbol \delta}

\newcommand{\boldgamma}{\boldsymbol \gamma}
\newcommand{\boldlambda}{\boldsymbol \lambda}

\newcommand{\boldphi}{\boldsymbol \phi}

\newcommand{\boldsigma}{\boldsymbol \sigma}

\newcommand{\boldtheta}{\boldsymbol \theta}
\newcommand{\boldvarphi}{\boldsymbol \varphi}

\newcommand{\bigo}[1]{O\left({#1}\right)}
\newcommand{\smallo}[1]{o\left({#1}\right)}

\papertype{Original Article}
\paperfield{Journal Section}

\title{Nonparametric Independent Component Analysis for the Sources with Mixed Spectra}

\abbrevs{ICA, independent component analysis}

\author[1]{Seonjoo Lee, PhD}
\author[2]{Haipeng Shen, PhD}
\author[3]{Young K. Truong, PhD}

\affil[1]{Mental Health Data Science, The Research Foundation in Mental Hygen Inc. and 
Department of Biostatistics and Psychiatry, Columbia University, U.S.A.}
\affil[2]{Innovation and Information Management, Faculty of Business and Economics, University of Hong Kong, Hong Kong, China}
\affil[3]{Department of Biostatistics, University of North Carolina at Chapel hill, Chapel hill, North Carolina, U.S.A.}

\corraddress{Seonjoo Lee PhD, Mental Health Data Science, The Research Foundation in Mental Hygen Inc. and 
Department of Biostatistics and Psychiatry, Columbia University, U.S.A}
\corremail{seonjoo.lee@nyspi.columbia.edu}

\fundinginfo{Lee’s work is partially supported by NIH grant R01AG062578. Shen’s work is partially supported by HKSAR CRF grant C7162-20G and HKU BRC grant.}

\runningauthor{Lee et al.}

\begin{document}
\maketitle
\begin{abstract}
Independent component analysis (ICA) is a blind source separation method to recover source signals of interest from their mixtures. Most existing ICA procedures assume independent sampling. Second-order-statistics-based source separation methods have been developed based on parametric time series models for the mixtures from the autocorrelated sources. However, the second-order-statistics-based methods cannot separate the sources accurately when the sources have temporal autocorrelations with mixed spectra. To address this issue, we propose a new ICA method by estimating spectral density functions and line spectra of the source signals using cubic splines and indicator functions, respectively. The mixed spectra and the mixing matrix are estimated by maximizing the Whittle likelihood function. We illustrate the performance of the proposed method through simulation experiments and an EEG data application. The numerical results indicate that our approach outperforms existing ICA methods, including SOBI algorithms. In addition, we investigate the asymptotic behavior of the proposed method.

\end{abstract}

\section{Introduction}\label{sec:int}
%

Independent component analysis (ICA) is a popular blind source separation method. A typical instantaneous ICA expresses a set of observed mixed signals as linear combinations of independent latent sources (or components):
\begin{equation}\label{eq:intro:ICA}
        \mb{X}_{M \times T}= \mb{A}_{M \times M}\mb{S}_{M \times T},
\end{equation}
where $\mb{X}$ is an $M\times T$ observed matrix of the $M$ mixed signals, $\mb{A}$ is a non-random mixing matrix, and $\mb{S}$ is the matrix of \textit{independent} source signals. 
Under appropriate conditions on $\mb{A}$, the source can be recovered as
\begin{equation} \label{eq:intro:ICAinv}
\mb{S}=\mb{W}\mb{X},\qquad \mb{W}=\mb{A}^{-1},
\end{equation}
by estimating the unmixing matrix $\mb{W}$, which maximizes independence criteria of the estimated sources.

Extensive literature reviews about methodological development and applications for ICA can be found in~\cite{hyvarinen2001ica},~\cite{hastie2009elements},~\cite{cichocki2009nonnegative} and~\cite{comon2010handbook}. More specifically, it has many important applications in
acoustic signal processing ~\citep{hyvarinen2001ica}; finance~\citep{back1997first}; medical image analysis, such as functional magnetic resonance imaging (fMRI)~\citep{calhoun2009rgi}, electroencephalography (EEG), and magnetoencephalography (MEG)~\citep{makeig2009erp}; and system monitoring~\citep{ge2007process}.

Most early works are based on information theory, including JADE~\citep{cardoso1993blind}, Infomax \citep{bell1995ima,lee1999independent}, and fastICA~\citep{hyvarinen2001ica}. Later works are based on maximum likelihood using nonparametric density estimation~\citep{vlassis2001esa, chen2006eic, bach2003kic, boscolo2004ica, chen2006fkd,hastie2003ica,kawaguchi2007sic}. Recent works attempt to directly measure independence using characteristic functions \citep{eriksson2003cfb,chen2005cic,matteson2011independent} and distance covariance \citep{matteson2017independent}.  When the sources are autocorrelated, second-order-statistics-based methods have been developed. Examples include joint-diagonalization of the covariance matrix and several autocovariance matrices at different lags \citep{belouchrani1997blind,miettinen2014deflation} and Whittle likelihood-based methods \citep{pham1997bso,lee2011ica,lee2020sampling}. 

In biomedical applications like EEG, the sources possibly include sinusoidal signals with line spectra (or atoms) in the spectral domain. Although the ARMA model or autocovariance can approximate sharp line spectra, the estimation is often biased due to the Dirichlet kernel effect. To address this gap, we propose a new ICA method by employing the flexible source models and log-spline density estimation procedure to the ICA. \cite{kooperberg1995lep} proposed a flexible model for the sources with mixed spectra; that is, the sources consist of sinusoidal signals and stochastic signals. The log-spline spectral density estimation procedure estimates the spectral density function, which is determined by the stochastic signal, with B-spline basis functions, and detects line spectra, which are determined by sinusoidal signals, based on Fourier frequencies. We take an iterative estimation procedure to estimate the unmixing matrix $\mb{W}$ and the spectral densities of the sources. In spectral density estimation, the numbers and locations of the knots of B-splines and the atoms are determined using Bayesian Information Criteria (BIC). Our simulation studies show better performance over the currently existing ICA algorithms. Finally, we investigated the asymptotic properties of the proposed method.

The rest of the paper is laid out as follows: Section \ref{sec:ica} reviews the Whittle likelihood-based ICA for autocorrelated sources and source models with possibly mixed spectra followed by the new ICA algorithm via nonparametric spectral density estimation (cICA-LSP). We show by intensive simulation studies that our newly proposed method performs better than other existing ICA algorithms in Section \ref{sec:sim}. In simulation studies, we also examine the performance of our new approach using real sound data. In Section \ref{lspec:sec:eeg}, we also analyze EEG data. In Section \ref{lspec:sec:theory}, asymptotic properties of cICA-LSP are presented.


\section{Independent Component Analysis via Nonparametric Spectral Density Estimation}\label{sec:ica}

This section proposes a new ICA procedure via maximizing the Whittle likelihood with the nonparametric spectral density estimation. First, we review the Whittle likelihood-based ICA method in Section~\ref{sec:whittle}, followed by a new source model and nonparametric spectral density estimation in Section~\ref{sec:icaintro}. Finally, Section ~\ref{sec:algorithm} presents the details of the cICA-LSP algorithm.

\subsection{Review of the Whittle likelihood-based ICA}\label{sec:whittle}
In this section, we review the Whittle likelihood-based ICA method proposed by \cite{lee2011ica} for temporally autocorrelated sources. For an observed $M$-dimensional vector process $\mb{X}(t) = (X_{1}(t),$ $ \ldots, X_{M}(t))^{\top}$ with zero mean and the covariance function $\boldgamma_{X}(u) = \Cov(\mb{X}(t), \mb{X}(t+u))$ and the $M\times M$ spectral density matrix $\mb{F_{X}}(\cdot)$. The discrete Fourier transform (DFT) of  $\mb{X}$ is given by  
\begin{equation*}
\mb{d}(\rad_k, \mb{X}) 
= \sum_{k=0}^{T-1}\mb{X}(t) \exp\{-i \rad_k t\}, \quad \rad_k = \frac{2\pi k}{T}, \quad k \in \{0,\ldots,T-1\}, 
      \end{equation*} 
and the second order periodogram by
$\PER(\rad_k,\mb{X})=\frac{1}{2 \pi T}{\mb{d}}(\rad_k, \mb{X}){\mb{d}}^*(\rad_k,\mb{X})$,
where $\mb{d}^{*}$ is the conjugate transpose of the vector $\mb{d}$. 
The spectral density matrix 
$\mb{F_S}$, the DFT and the second order periodogram of the source signals 
$\mb{S}(t) = (S_1(t),\ldots,S_M(t))^{\top}$, $t \in \{0, 1, \ldots, T-1\}$, 
are defined similarly. Note that the source spectral density matrix is diagonal $\mb{F_{S}}=\mr{diag}(f_{1}, \ldots, f_{M})$, 
where $f_{j},~ j \in \{1,\ldots, M\}$ is the spectral density of the $j$th source because the sources are mutually independent.

By the independence of the sources, the ICA model~\refp{eq:intro:ICAinv}, 
and the asymptotic properties of the DFT, we derive the Whittle log-likelihood~\citep{dzhaparidze1986peh}:

\begin{equation}\label{eq:whittle:clICA}
\begin{split}
  \mathcal{L}(\mb{W}, \mb{F_{S}}; \mb{X})
	&= -\frac{1}{2T} \sum_{j=1}^{M} \sum_{k=0}^{T-1} 
	\left(\frac{ \mb{e}_{j}^{\top}\mb{W}\PER(\rad_{k},\mb{X}) \mb{W}^{\top}
	\mb{e}_{j}}{f_{j}(\rad_{k})} + \log{f_{j}(\rad_{k})} \right)+ \log|\mr{det}(\mb{W})|,
\end{split}
\end{equation}
where $\mb{e}_{j}=(0,0,\ldots,1,0,\ldots,0)^{\top}$ 
with the $j$th entry being 1. 

Note that \cite{lee2011ica} parametrized the source spectra using autocorrelation (AR) models, and the unmixing matrix $\mb{W}$ was estimated by maximizing \refp{eq:whittle:clICA}. The method has shown promising results through simulation studies and was applied to fMRI data to identify brain-functional areas. For biomedical applications, however, some sources may have more complex structures. In EEG experiments, for instance, ICA is often applied to reduce noise artifacts or sometimes to detect neuronal signals. Some potential sources of interest are known to have a speciﬁc frequency range, such as the alpha rhythm (a neural oscillation) with a frequency range between 8–12 Hz. In the spectral domain of time series analysis, such a source has a continuous spectral density (for the background noise) mixed with a discrete real biological component called the line spectrum. The AR model is generally ﬂexible to approximate a wide variety of continuous spectral density functions, while some harmonic processes are known to be useful for modeling line spectra. In this paper, we employ a nonparametric spectral density estimation technique to improve the flexibility in simultaneously estimating the continuous and line spectra of the sources. 

\subsection{Source Model}\label{sec:icaintro}
To model source spectral density functions, we consider sources with mixed spectra, studied broadly in literature. In this section, we review a mixed spectral density model proposed by \cite{kooperberg1995lep}.

Consider a univariate second order stationary process $\{S_j(t)\}$ with mean zero and covariance function $\gamma_j(u) = \Cov(S_j(t),S_j(t+u))$ for $j \in \{1,\ldots,M\}$. Assume that each source time series has the form
\begin{equation}\label{eq:lspec:mixedmodel} 
  S_j(t) =\sum_{p=1}^{P_j} R_{jp} \cos(t \omega_{jp} + \phi_{jp} ) + Y_j(t),\quad t \in \{0,\ldots,T-1\},
\end{equation}
where $0<\omega_{jp} \leq \pi$; $\phi_{jp}$ are independent and 
uniformly distributed on $[-\pi, \pi]$; $R_{jp},~p=1,\ldots,P_j,$ are 
independent, non-negative random variables such that $R_{jp}^{2}$ has 
positive mean $4\rho_{jp}$; and $Y_j(t)$ is a second-order 
stationary time series with mean zero and spectral density function $f_j^c$, and is independent of $R_{jp}$ and $\phi_{jp}$. 

Model \refp{eq:lspec:mixedmodel} satisfies the weak stationary conditions and is very flexible, allowing varying phases and intensities of deterministic terms. In addition, we note that~\refp{eq:lspec:mixedmodel} is a very practical model for many applications, such as fMRI or EEG data analysis, because most brain-function-related or physiological signals have mixed spectral densities. Then, the spectral density distribution of the time series $S_j(t)$ is given by 
\begin{equation}
	  F_j(\rad) = \int_{-\pi}^{\rad} f_j^{c}(u) \text{d}u + 
	  \sum_{u \leq \rad} f_j^{d}(u), \quad -\pi \leq \rad \leq \pi,
\end{equation}
where its line spectrum is given by 
\begin{equation}
	  f_j^{d}(\rad)=
	  \begin{cases}
	    \rho_{jp} & \text{if } \rad = \pm \lambda_{jp}\\
	    0 & \text{ otherwise.}
	  \end{cases}
	\end{equation}
Note that $f_j^{c}$ and $f_j^{d}$ are symmetric about zero and periodic 
with period $2\pi$. 

The periodogram of each source is given by
\begin{equation*}
  \per(\rad,S_j) = \frac{1}{2\pi T} \left|\sum_{t=0}^{T-1} e^{-i\rad t}S_j(t) \right|^2, \quad -\pi \le \rad \le \pi.
\end{equation*}
Denote the mean spectral density function of each source as $f_j=f_j^c+\frac{T}{2\pi}f_j^d$. The periodogram, $\per(\frac{2\pi k}{T},S_j)$, asymptotically follows $f_j(\frac{2\pi k}{T}) E_j$, where $E_j~\sim \text{Exp}(1)$ if $k<T/2$ and $E_j\sim\chi_1^2$ if $T$ is even and $k=T/2$, and are asymptotically independent~\citep{brillinger2001tsd}.

Below, we describe how to model the mean spectral density functions using a nonparametric scheme. There have been many nonparametric spectral density estimation methods. Examples include smoothing the periodogram~\citep{brillinger2001tsd}, minimizing least-square using smoothing splines with cross-validation~\citep{wahba1980asl}, maximizing the Whittle likelihood via sieve estimation~\citep{chow1985sms}, bootstrapping~\citep{franke1992bootstrapping}, penalized Whittle likelihood~\citep{pawitan1994nsd}, log-spline with knot selection based on BIC~\citep{kooperberg1995lep}, and local smoothing with adaptive bandwidth selection~\citep{fan1998automatic}. Among all, we employ the log-spline spectral density estimation procedure proposed by~\cite{kooperberg1995lep} due to the ability to detect deterministic periodic signals and the convenience of combining the ICA algorithm. 

For each $j  \in \{1,\ldots,M\}$, write the logarithm of the mean spectral density function as $\varphi_{j} = \log{f_{j}}$. We use a combination of B-splines and dirac delta functions to model the spectral density functions $f_j^c$ and the line spectra $f_j^d$, respectively. 

First, we model the spectral density functions $f_j^c$. Given the positive integer $K_j^c$ and the sequence $t_{j1}, \ldots, t_{jK_j^c}$, with $0 \leq t_{j1} < \cdots < t_{jK_j^c} \leq \pi$, let $S_{T}^{j,c}$ denote the space of cubic polynomial splines $s$ on the interval $[0,t_{j1}],\ldots,[t_{jK_j^c},\pi]$ with the first derivative of $s$ being zero at $0$ and $\pi$, the third derivative of $s$ being zero at $0$ unless $t_{j1}=0$, and the third derivative of $s$ being zero unless $t_{jK_j^c}=\pi$. Let $B_{jk}, 1 \le k \le K_j^c$, denote the usual basis of $S_{T}^{j,c}$ consisting of B-splines for spline $g_{j}^c$. Since the spectral density function $f_{j}^c$ is symmetric about zero, and is periodic with the period $2\pi$, we have that $f_{j}^{c'}(0) = f_{j}^{c'''}(0) = f_{j}^{c'}(\pi) = f_{j}^{c'''}(\pi)=0$ and $\varphi_{j}^{c'}(0) = \varphi_{j}^{c'''}(0) = \varphi_{j}^{c'}(\pi) = \varphi_{j}^{c'''}(\pi)=0$. Here, splines $g_j^c$ on $[0, \pi]$ such that $g_{j}^{c'}(0) = g_{j}^{c'''}(0) = g_{j}^{c'}(\pi) = g_{j}^{c'''}(\pi)=0$ are considered to model the logarithm of the spectral density function $\varphi_{j}$. 

To model line spectrum, we consider the space of dirac delta functions. Given the positive integer $K_j^d$ and the increasing sequence $a_{1}, \ldots,a_{K_j^d}\in\left\{\frac{2\pi j}{T}: 1 \leq j\leq\frac{T}{2}\right\}$, let $S_T^{j,d}$ be the $K_j^d$-dimensional space of nonnegative function $s$ on $[0, \pi]$ such that $s=0$ except $a_{1}, \ldots, a_{K_j^d}$. Set $B_{j k+K_j^c}(\omega) = \delta_{a_{j}}(\omega), 1\leq k \leq K_j^d$. Then $B_{1+K_j^c}, \ldots, B_{K_j^d+K_j^c}$ is a B-spline basis of $S_T^{j,d}$. 

Finally, we combine two procedures described above together to estimate the spectral density. Let $S_T^j$ be the space spanned by $B_{j1}, \ldots, B_{jK_j}$, where $K_j=K_j^c+K_j^d$. Set 
\begin{equation*}\begin{split}
g_j^c &= \beta_{j1}B_{j1} + \cdots +\beta_{jK_j^c}B_{jK_j^c},\\
g_j^d &= \beta_{j K_j^c+1}B_{j K_j^c+1} + \cdots +\beta_{j K_j}B_{jK_j},
\end{split}\end{equation*}
with $\beta_{jK_j^c+1}, \ldots, \beta_{jK_j} \geq 0$, and 
$$
g_j(\cdot;\boldbeta) = g_j^c(\cdot;\boldbeta_{c})  + g_j^d(\cdot;\boldbeta_{d})
= \beta_{j1} B_{j1}(\cdot) + \cdots + \beta_{jK_j} B_{jK_{j}}(\cdot) = \boldbeta_j^{\top}\mb{B}_j(\cdot),
$$
with $\boldbeta_j= (\beta_{j1}, \ldots, \beta_{jK_{j}})^{\top},~j=1,\ldots,M$, 
and $\sum_{j=1}^{M}K_{j}= K$ so as to 
$\boldbeta = (\boldbeta_{1}^{\top}, \ldots, \boldbeta_{M}^{\top})^{\top}$ 
constrained to lie in the subspace $\Omega$ of $\mathbb{R}^{K}$ given by
\begin{multline*}\label{eq:lspecica:st}
 	 \Omega =\left\{ \boldbeta = (\boldbeta_{j}^{\top}, \ldots, 
	 \boldbeta_{M}^{\top})^{\top} = ( \beta_{j1}, \ldots,
     	\beta_{MK_{M}})^{\top} \in \mathbb{R}^{K} : \right.\\
	\text{for each }j,~g_{j}^{'}(0) = g_{j}^{'''}(0) = g_{j}^{'}(\pi) =
     	g_{j}^{'''}(\pi)=0, \\
	\left.\text{where } g_{j} =\beta_{j1} B_{j1}(\cdot) + 
	\cdots + \beta_{jK_{j}}B_{K_{j}}(\cdot)\right\}.
\end{multline*}

The Whittle loglikelihood is rewritten in terms of the unknown parameters $(\mb{W}, \boldbeta)$ as
\begin{equation}\label{eq:lspecica:wlICA_nonpara}
\begin{split}
  \mathcal{L}_T(\mb{W}, \boldbeta; \mb{X})
  &= -\frac{1}{2T} \sum_{j=1}^{M}\sum_{k=0}^{T-1} 
  \left(\frac{ \mb{e}_{j}^{\top}\mb{W} \PER(\rad_{k},\mb{X}) \mb{W}^{\top} \mb{e}_{j}}{ \exp\{{g_{j}(\rad_{k};\boldbeta_{j})}\}} 
  + g_{j}(\rad_{k};\boldbeta_{j})\right) + \log{|\det{\mb{W}}|}\\
   &= -\frac{1}{2T}\sum_{j=1}^{M}\sum_{k=0}^{T-1} 
   \left(\frac{ \mb{e}_{j}^{\top}\mb{W} \PER(\rad_{k},\mb{X}) \mb{W}^{\top} \mb{e}_{j}}{ \exp\{\boldbeta_j^{\top}\mb{B}_j(\rad_k)\}} 
  +  \boldbeta_j^{\top}\mb{B}_j(\rad_k)\right) + \log{|\det{\mb{W}}|}.
\end{split}
\end{equation}
Note that when the data are spatially whitened (or sphered), we have $\log{|\det{\mb{W}}|} = 0$ by the orthonormal condition. Once the Whittle log-likelihood is formulated as above, we can obtain the estimates for the unmixing matrix $\mb{W}$ and the estimates of the power spectra by maximizing \refp{eq:whittle:clICA}.

\textit{Remarks.} 
1.  As a reminder, the approach of~\cite{pham1997bso} assumed that $f_{j}$ is known, and~\cite{lee2011ica} used the ARMA model to estimate spectral densities. In this paper, we estimate spectral densities using an adaptive nonparametric method that requires weaker assumptions that previous methods and is flexible for mixed spectra.

2. In spectral density estimation, if we fix knots, spectral density estimation becomes a fully parametric approach. In our procedure, we select the number of knots. For detail on the knot selection procedure, see \cite{kooperberg1995lep}.

\subsection{Algorithm}\label{sec:algorithm}

Many ICA algorithms apply (spatial) pre-whitening as a preprocessing step due to many advantages of optimization on the Stiefel manifold~\citep{edelman1998geometry, amari1999natural, douglas2002self,plumbley2004lie, ye2006monotonic}. Let $\boldsymbol{\Sigma}_X=\Cov \mb{X}$ and $\mb{\widetilde{X}}=\boldsymbol{\Sigma}_{\mb{X}}^{-1/2}\mb{X}$. 
Then, $\Cov \widetilde{\mb{X}} = \mb{I}_{M\times M}$ and \refp{eq:intro:ICAinv} is equivalent to $\mb{S}=\mb{W} \boldsymbol{\Sigma}_{\mb{X}}^{1/2}\widetilde{\mb{X}}$. If we assume $\Cov \mb{S}=\mb{I}_{M\times M}$, we have $\mb{O}=\mb{W}  \boldsymbol{\Sigma}_{\mb{X}}^{1/2}$, be an orthonormal matrix. Now the problem is to solve 
$\mb{\widetilde{X}=OS}$,
restricting the matrix $\mb{O}$ on the orthonormal matrices. The covariance matrix $ \boldsymbol{\Sigma}_{\mb{X}}$ can be estimated by the sample covariance matrix of $\mb{S}$, $\widehat{\boldsymbol{\Sigma}}_{\mb{X}}=\frac{1}{T} \sum_{t=0}^{T-1} \left(\mb{X}(t)-\bar{\mb{X}}\right)\left(\mb{X}(t)-\bar{\mb{X}}\right)^{\top}$, where $\bar{\mb{X}}=\frac{1}{T}\sum_{t=0}^{T-1} \mb{X}(t)$. The prewhitened ICA algorithms first estimates an orthogonal unmixing matrix, $\mb{\widehat{O}}$, by solving \refp{eq:lspecica:wlICA_nonpara}, and then estimate the unmixing matrix by 
$\widehat{\mb{W}}=\widehat{\mb{O}}\widehat{ \boldsymbol{\Sigma}}_{\mb{X}}^{-1/2}$. 

To incorporate the orthonormal constraints, we propose to use Lagrange multiplier. More specifically, we consider minimizing the following penalized negative Whittle log-likelihood,    
    \begin{equation}\label{eq:lspecica:lagrange}
      F(\mb{O} ,\boldlambda, \boldbeta) = -\mathcal{L}_T(\mb{O},\boldbeta;\widetilde{\mb{Y}})
      +{\boldlambda}^{\top} \mb{C},
    \end{equation}
where ${\boldlambda} = (\lambda_{1},\ldots,\lambda_{M(M+1)/2})^{\top}$ is 
the Lagrange parameter vector, and $\mb{C}$ is a $M(M+1)/2$-dimensional vector 
with the element being $C_{(j-1)M+k} = (\mb{OO}^{\top}-\mb{I}_{M})_{jk},
~j=1,\ldots,M, k =1,\ldots,j$. Note that we need only $M(M+1)/2$ constraint 
functions since $\mb{OO}^{\top}-\mb{I}_{M}$ is symmetric.

In practice, the locations and the number of knots of the spectral densities are not given, so the knot selection procedure is necessary. However, minimizing~\refp{eq:lspecica:lagrange} with the knot selection procedure and the parameters at the same time is computationally expensive. Henceforth, we iteratively estimate the unmixing matrix and the spectral density parameters. In each iteration, given $\widetilde{\mb{O}}$, we estimate $\boldbeta_{j}$ maximizing the Whittle log-likelihood~\refp{eq:lspecica:wlICA_nonpara}. We then select the "optimal'' knots using model selection criterion such as Bayesian information criterion~(BIC)~\citep{schwarz1978edm} as described in Section \ref{sec:icaintro}.
    
Next, given the estimate $\tilde{\boldbeta}$, we obtain the updated estimate of unmixing matrix $\mb{O}$ by minimizing the 
penalized criteria \refp{eq:lspecica:lagrange}. Since~\refp{eq:lspecica:lagrange} is nonlinear, we used Newton-Raphson method with Lagrange multiplier. To begin with, we find the first and second derivatives of~\refp{eq:lspecica:lagrange} 
with respect to $\mb{O}$ and $\boldlambda$ and denote them as 
\[\nabla F(\mb{O},\boldlambda;\boldbeta) 
=\begin{pmatrix}
\nabla_{\mb{O}}F\\\nabla_{\boldlambda}F
\end{pmatrix}
=\begin{pmatrix}
  \frac{\partial F}{\partial \vect{\mb{O}}}\\ 
  \frac{\partial F}{\partial \boldlambda }
\end{pmatrix} \] and 
\[\nabla^2 F(\mb{O},\boldlambda;\boldbeta)
= \begin{pmatrix}
  \mb{H}_1 & \mb{H}_2\\
  \mb{H}_2^{\top} & \mb{0}
\end{pmatrix},\]
where 
\[[\mb{H}_1]_{jk}=\frac{\partial^2 F}{\partial \vect{\mb{O}}_j \partial\vect{\mb{O}}_k},~[\mb{H}_2]_{jk}=\frac{\partial^2 F}{\partial \vect{\mb{O}}_j \partial\boldlambda_k}.\]   

Then we obtain the one-step Newton-Raphson update for $\mb{W}$ and $\boldlambda$ as

\begin{equation}\begin{split}
  \mb{O}^{(new)}
  &=\mb{O}^{(old)} - \mb{H_{1}}^{-1}
  \left(\mb{I}-\mb{H}_2\left(\mb{H}_2^{\top}\mb{H}_1^{-1}\mb{H}_2\right)^{-1}\mb{H}_2^{\top}\mb{H}_1^{-1}\right) \nabla_{\mb{O}}F \\
  &\quad - \mb{H}_1^{-1}\mb{H}_2\left(\mb{H}_2^{\top}\mb{H}_1^{-1}\mb{H}_2\right)^{-1}\nabla_{\boldlambda}F,\\
  \boldlambda^{(new)}
  &=\boldlambda^{(old)} - \left(\mb{H}_2^{\top}\mb{H}_1^{-1}\mb{H}_2\right)^{-1}\mb{H}_2^{\top}\mb{H}_{1}^{-1}\nabla_{\mb{O}}F + \left(\mb{H}_2^{\top}\mb{H}_1^{-1}\mb{H}_2\right)^{-1}\nabla_{\boldlambda}F ,
\end{split}\end{equation}
where $\mb{H}_1$, $\mb{H}_2$, $\nabla_{\mb{O}}F$, and $\nabla_{\boldlambda}F$ depend on $\mb{O}^{(old)}$ and $\boldlambda^{(old)}$.
    
The procedure to update $(\mb{W}, \boldlambda)$ and $(\boldphi, \boldsigma^{2})$ 
will be alternated until convergence. Since ICA methods have permutation and scale ambiguity problem,  
we use Amari's Distance \citep{amari1996nla} as the convergence criterion defined as 
 \begin{equation}\label{eq:lspecica:amarisdist}
      d_{Amari}(\mb{M}_{1},\mb{M}_{2}) = \frac{1}{M} \sum_{i=1}^{M}\left(
        \frac{\sum_{j=1}^{M}|a_{ij}|}{\max_{j}|a_{ij}|} -1 \right) +
        \frac{1}{M} \sum_{j=1}^{M}\left(
        \frac{\sum_{i=1}^{M}|a_{ij}|}{\mathop{\rm \max}_{i}|a_{ij}|} -1 \right),
    \end{equation}
    where $\mb{M}_{1}$ and $\mb{M}_{2}$ are $M \times M$ matrices, $\mb{M}_{2}$ is invertible and $a_{ij}$ is the $ij$th element of $\mb{M}_{1}\mb{M}_{2}^{-1}$. When the Amari's distance is less than some threshold, the iteration stops. We name this method as cICA-LSP. The algorithm is summarized as follows:\\

\centerline{\fbox{\bf The cICA-LSP Algorithm}}\vskip -0.05in
\rule{6.3in}{.1mm}
    \begin{description}
      \item[\bfseries{\em Pre-whitening}] $\widetilde{\mb{X}} = \widehat{\boldsymbol{\Sigma}}_X^{-1/2}\mb{X}$.
      \item[\bfseries{\em Initialize}] $\mb{O}^{(old)}, \boldlambda^{(old)}$.
      \item[\bfseries{\em While}] the convergence criterion is unsatisfied,	\begin{enumerate}
	  \item For given $\mb{O}^{(old)}$, estimate the sources: $\widehat{\mb{S}} = \mb{O}^{(old)}\widetilde{\mb{X}}$. For each $j=1,\ldots,M$,
		\begin{enumerate}		    
		  \item Select the location and the number of knots and their locations.
		  \item Estimate $\tilde{\boldbeta}_{j}$.
		\end{enumerate}
	  \item For given $\tilde{\boldbeta}$, update $\mb{O}^{(old)}$ and $\boldlambda^{(old)}$ as 
	   \begin{equation*}\begin{split}
  \mb{O}^{(new)}
  &=\mb{O}^{(old)} - \mb{H_{1}}^{-1}
  \left(\mb{I}-\mb{H}_2\left(\mb{H}_2^{\top}\mb{H}_1^{-1}\mb{H}_2\right)^{-1}\mb{H}_2^{\top}\mb{H}_1^{-1}\right) \nabla_{\mb{O}}F \\
  &\quad - \mb{H}_1^{-1}\mb{H}_2\left(\mb{H}_2^{\top}\mb{H}_1^{-1}\mb{H}_2\right)^{-1}\nabla_{\boldlambda}F,\\
  \boldlambda^{(new)}
  &=\boldlambda^{(old)} - \left(\mb{H}_2^{\top}\mb{H}_1^{-1}\mb{H}_2\right)^{-1}\mb{H}_2^{\top}\mb{H}_{1}^{-1}\nabla_{\mb{O}}F + \left(\mb{H}_2^{\top}\mb{H}_1^{-1}\mb{H}_2\right)^{-1}\nabla_{\boldlambda}F ,
\end{split}\end{equation*}
where $\mb{H}_1$, $\mb{H}_2$, $\nabla_{\mb{O}}F$, and $\nabla_{\boldlambda}F$ depend on $\mb{O}^{(old)}$ and $\boldlambda^{(old)}$. 
\item $\mb{O}^{(old)}\leftarrow\mb{O}^{(new)},~\boldlambda^{(old)}\leftarrow\boldlambda^{(new)}$
\end{enumerate}
      \item[\bfseries{\em end}]
      \item[\bfseries{\em Final} $\widehat{\mb{W}} = \mb{O}\widehat{\boldsymbol{\Sigma}}_X^{-1/2}$. ]
\end{description}
\vspace*{-2mm} \rule{6.3in}{.1mm}

\section{Simulation Studies}\label{sec:sim}
The first experiment in Section \ref{sec:sim1} is designed to illustrate the performance of the newly proposed algorithm (cICA-LSP). In Section \ref{sec:sound}, we apply the proposed method to real audio signals obtained from three music sources and two randomly generated noise signals. 
\subsection{Simulation study I: sources with mixed spectra-Fourier frequency atoms}\label{sec:sim1}

According to the ICA model \refp{eq:intro:ICA}, we first generate 
the source matrix $\mb{S}$, consisting of the four ($M=4$) stationary time series with two different sample sizes ($T=512, 4096$). The four sources are 
generated from: 
\begin{enumerate}
  	\item $S_{1}(t)=2\sum_{j=1}^{3}\cos(\omega_{1j}t+\phi_{1j})+Y_{1}(t)$, $Y_{1}$ $\sim~ N(0,1)$;
	\item $S_{2}(t)=2\sum_{j=1}^{3}\cos(\omega_{2j}t+\phi_{2j})+Y_{2}(t)$, $Y_{2}$ $\sim~U(-\sqrt{3}, \sqrt{3})$;
	\item $S_{3}(t)=2\sum_{j=1}^{3}\cos(\omega_{3j}t+\phi_{3j})+Y_{3}(t)$,  $Y_{3}$: AR(1) with $t(3)/\sqrt{3}$;
	\item $S_{4}(t)=2\sum_{j=1}^{3}\cos(\omega_{4j}t+\phi_{4j})+Y_{4}(t)$,  $Y_{4}$: MA(1) with $N(0,1)$,
\end{enumerate}
where $\omega_{jk},~j=1,2,3,4,~k=1,2,3$, are 
chosen as follows:
$\omega_1 = \frac{2\pi}{128}(1,2,3)^{\top}$; 
$\omega_2 = \frac{2\pi}{512} + \frac{2\pi}{64}(1,2,3)^{\top} $; 
$\omega_3 = \frac{2\pi}{64}(1,2,3)^{\top}$; 
$\omega_4 = \frac{2\pi}{128} + \frac{2\pi}{64}(1,2,3)^{\top}$.

A $4\times 4$ full-rank mixing matrix $\mb{W}$ was randomly generated as 
\begin{equation*}
\mb{W}= 
\begin{pmatrix}
0.56 & 0.58 & -0.07 & 0.59\\
 -0.41 &  0.84 & 0.10 & 0.34\\
 -0.15 & 0.05 & 0.75 & -0.65\\
  0.53 & -0.83 & -0.08 & 0.13
\end{pmatrix}.
\end{equation*}

We also performed the simulation experiments for multiple randomly generated $\mb{W}$s to evaluate the uniform performance across the true unmixing matrix (Supplementary). The data matrix $\mb{X}$ is then obtained by multiplying $\mb{A}$ and $\mb{S}$: $\mb{X=AS}$. The simulation is replicated 100 times. The performance of cICA-LSP is compared to the existing popular methods, JADE~\citep{cardoso1993blind}, fastICA~\citep{hyvarinen2001ica}, JADE-SOBI ~\citep{belouchrani1997blind}, SOBIN, SOBIdefl \citep{miettinen2014deflation}, and  cICA-YW~\citep{lee2011ica}.

\begin{figure}
\centering
\includegraphics[width=\textwidth]{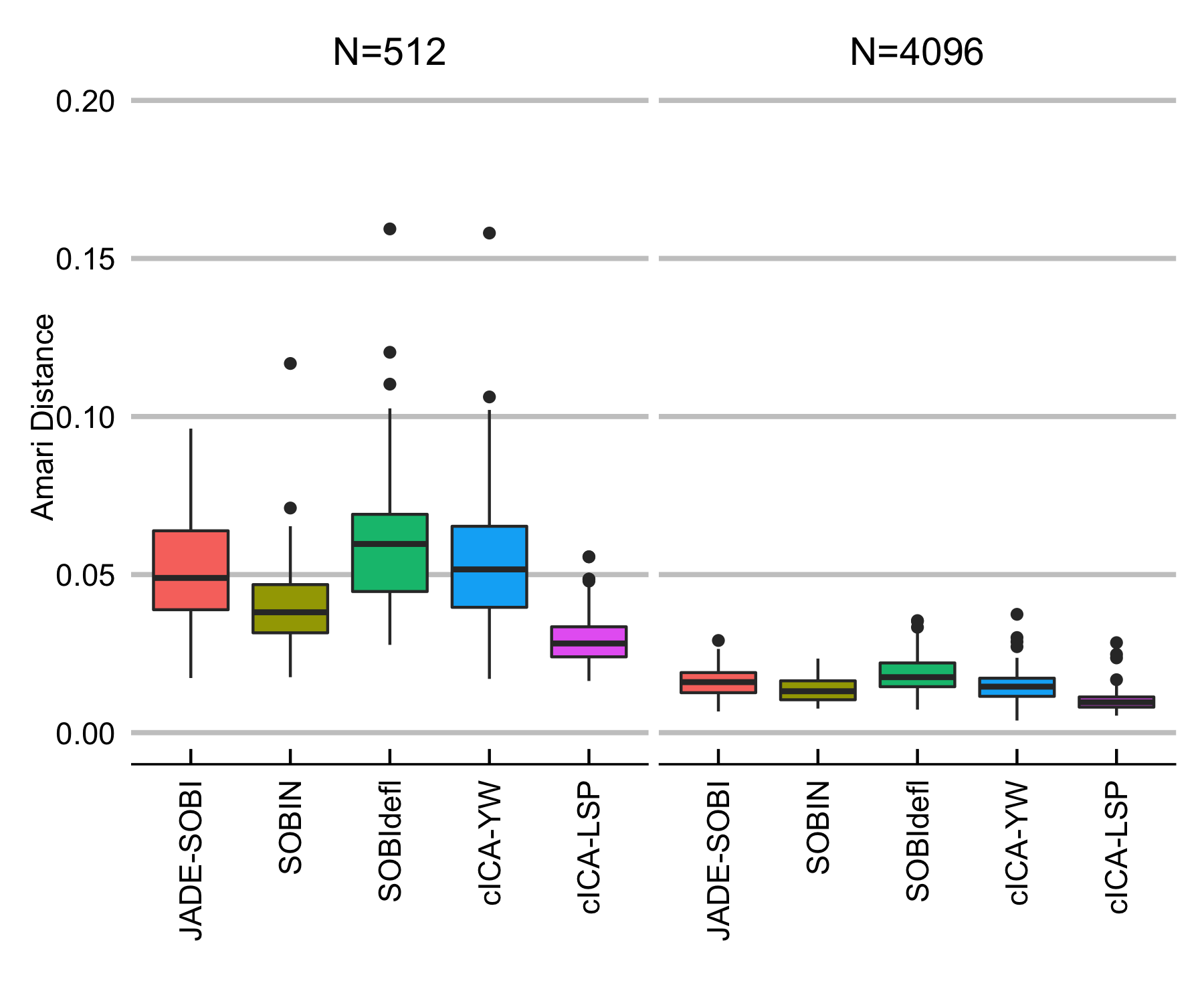}
\caption[Simulation Study I: Performance comparison for sources with mixed spectra of atoms at Fourier frequency]
{Simulation Study I: Performance comparisons for sources with mixed spectra of atoms at Fourier frequency. One hundred simulation runs are 
performed with different sample sizes ($512,~4096$). In each 
simulation run, $4$ sources with mixed spectra are generated and are 
mixed through a $4 \times 4$ mixing matrix. The boxplots show the Amari 
distance between the true unmixing matrix and the estimated unmixing matrix 
obtained by various ICA methods. The cICA-LSP provides more accurate estimates than the other existing 
methods.}\label{fig:sim:amari}
\end{figure}

The Amari distance \citep{amari1996nla} described in Section~\ref{sec:algorithm} is used as a performance comparison criterion between various ICA algorithms. Figure~\ref{fig:sim:amari} shows the boxplots of the Amari error for each method with different sample sizes. Since JADE and fastICA performed far worse than the methods that handle autocorrelations, we did not include those two methods here. The full simulation results are reported in the appendix (Figure~\ref{fig:sim:amarifull}). The cICA-LSP performed better than the existing methods across the different sample sizes. 

\subsection{Audio Sound Example}\label{sec:sound}
\begin{figure}
\begin{center}
\includegraphics[width=0.99\textwidth]{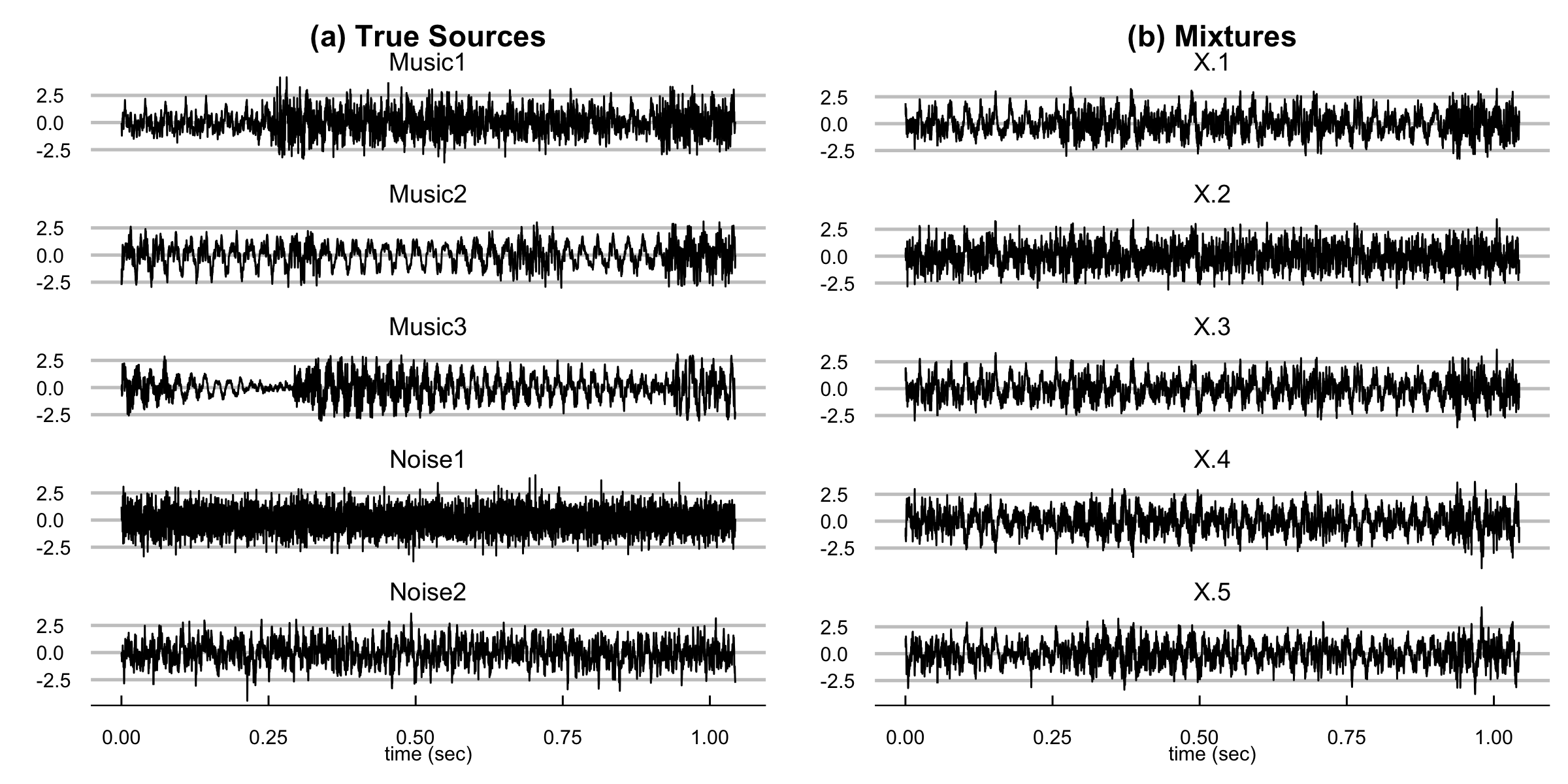}
\end{center}
\caption{The source and mixed signals. Three music sounds and two simulated noise artifacts are mixed with a randomly generated matrix.}\label{fig:sound:s}
\end{figure}

In this section, we apply the proposed method to a mixture of three music sounds and two simulated noise artifacts. Each music signal is extracted a length 4 seconds and sampled $11025Hz$ resulting 44100 time points. Two noise signals are generated as:
\begin{itemize}
  \item $N_1(t)=2\sin(0.036\pi t + 0.1234)+z_1(t)$, $z_1\sim N(0,1)$;
  \item $N_2(t)=2\cos(0.007\pi t)+ z_2(t)$, $z_2(t)\sim AR(0.8)$.
\end{itemize}
To generate noisier signals, we generated noise from the logistic distribution for $N_2$ \citep{chang1999composing}. The five signals are mixed with a randomly generated 5-by-5 matrix, and the mixed signals are separated using the cICA-LSP algorithm and compared with other ICA algorithms. Figure \ref{fig:sound:s} displays the true sources ($\mb{S}$) and the mixture ($\mb{X}$) in panels (a) and (b), respectively. 

\begin{figure}\begin{center}
\includegraphics[width=0.99\textwidth]{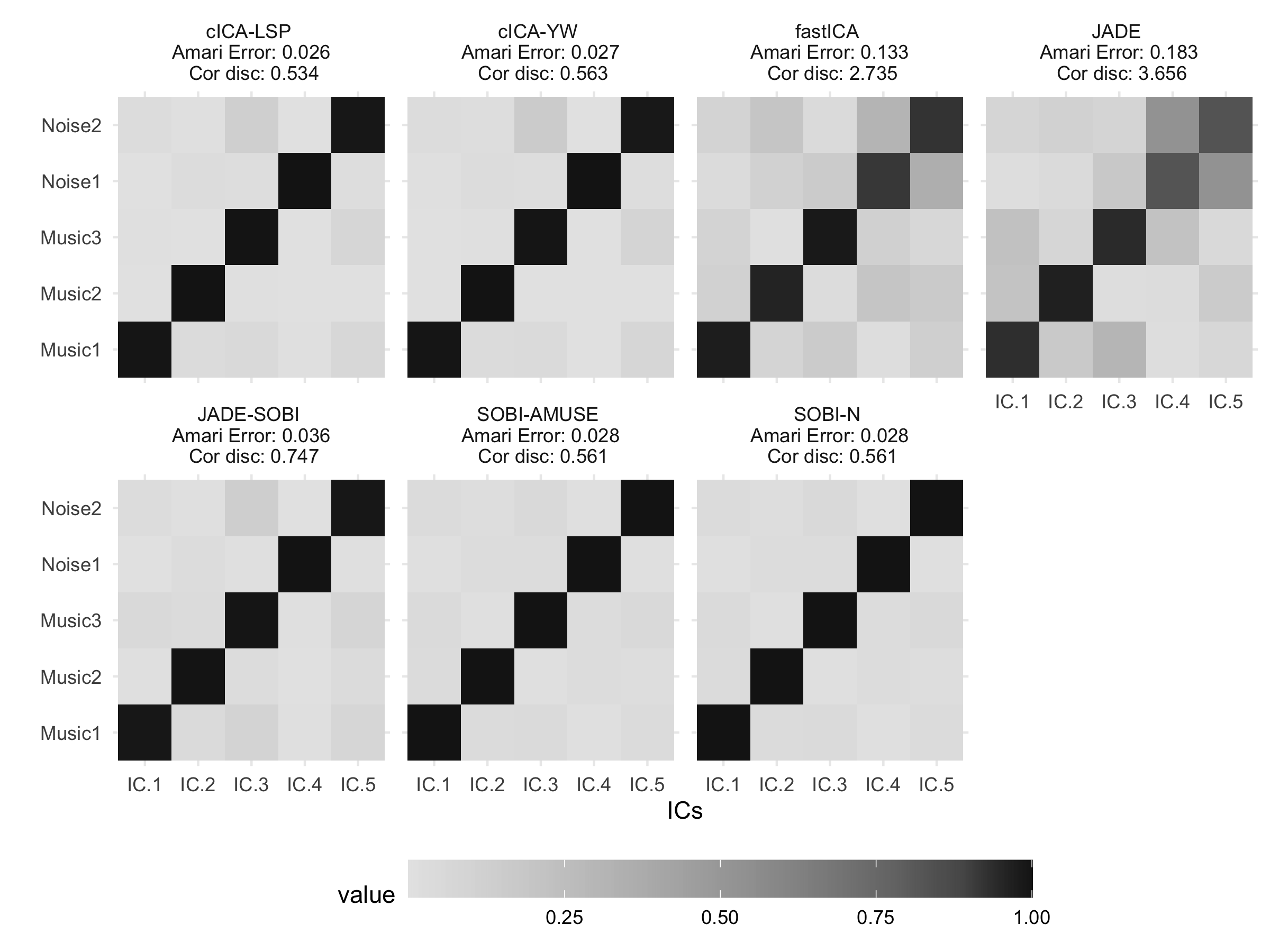}
\end{center}\caption{Each panel represents a five-by-five correlation matrix, and each entry is colored with a gray color scheme where a dark color represents a high correlation (1-black), and a bright color represents low correlation (0-light gray). Both cICA-LSP and cICA-YW perform better than the other two marginal density-based ICA algorithms in terms of the correlation matrix, and cICA-LSP has the smallest Amari error and correlation discrepancy.}
\label{fig:sound:correlation}
\end{figure}

To compare the performance of the ICA algorithms, we consider two criteria: correlations between the estimated sources and the true sources; Amari error between the true unmixing matrix and estimated matrices. Figure \ref{fig:sound:correlation} shows the correlation matrices obtained by seven ICA methods. Due to the sign and permutation ambiguity of ICA algorithms, we report absolute values of correlations and reordered the row of the matrix according to the highest correlation with the true sources. Each panel represents a five-by-five correlation matrix, and each entry is colored with a gray color scheme where a bright color represents a high correlation (1-white), and a dark color represents low correlation (0-black). Amari errors are reported under each correlation matrix. We also present the discrepancy between the recovered and true sources computed as the sum of the absolute value of the off-diagonal elements of the correlation matrix.  A lower discrepancy (Cor disc) indicates better performance. The cICA-LSP has the smallest Amari error and correlation discrepancy among the compared methods.

\section{EEG Data Example}\label{lspec:sec:eeg}

EEG signals often include artifacts that do not originate from the brain. The sources of artifacts include power line artifacts (50 or 60 Hz), muscle artifacts, and eye movement. The power line artifact can often be removed by low-pass filtering. Other artifacts can be separated from the brain signals through ICA. Example data is downloaded through the link that the EEGUtils package provides (https://craddm.github.io/eegUtils/articles/eegUtils.html). The data was recorded at 1024 Hz using a BioSemi ActiveTwo amplifier and active electrodes and was downsampled to 256 Hz. There were 64 electrodes positioned and named according to the 10-05 international system. Four additional electrodes (EXG1-EXG4) were placed around the eyes to record eye movements, and two further reference electrodes were placed on the left and right mastoids (EXG5 and EXG6). EXG7 and EXG8 are empty channels with no electrodes attached. 

As a part of preprocessing, the data was referenced to a common average calculated from all the electrodes after removing the empty channels (EXG7 and EXG8). For reference, the eye movement channels (EXG1-EXG4) were not included in the computing average. Then, we performed Finite Impulse Response (FIR) filtering with a high-pass filter at .1 Hz and a low-pass filter at 40 Hz. Data were epoched around the onset of a visual target on the left and right of fixation. We corrected the baseline using the average time points from -.1 to 0 seconds from the stimulus onsets and limited the range from -.1 to .4 seconds around the stimulus onsets. To run ICA, we included the 64 electrodes and did not include EXG1-EXG6 to focus on the neural signals. The parallel analysis \citep{dinno2009exploring} estimated the dimension of ICs as 29. The cICA-LSP and SOBI extracted 29 independent components (Figures \ref{fig:eeglsp} and \ref{fig:eegsobi} in Appendix). We identified the noise artifacts as the ICs that eye movement (EXG1-EXG4) or mastoid signals (EXG5-6) explained more than 13\% of the variance, which is equivalent to medium effect size (Cohen's $f^2$=0.15).

\begin{figure}\begin{center}
\includegraphics[width=0.8\textwidth]{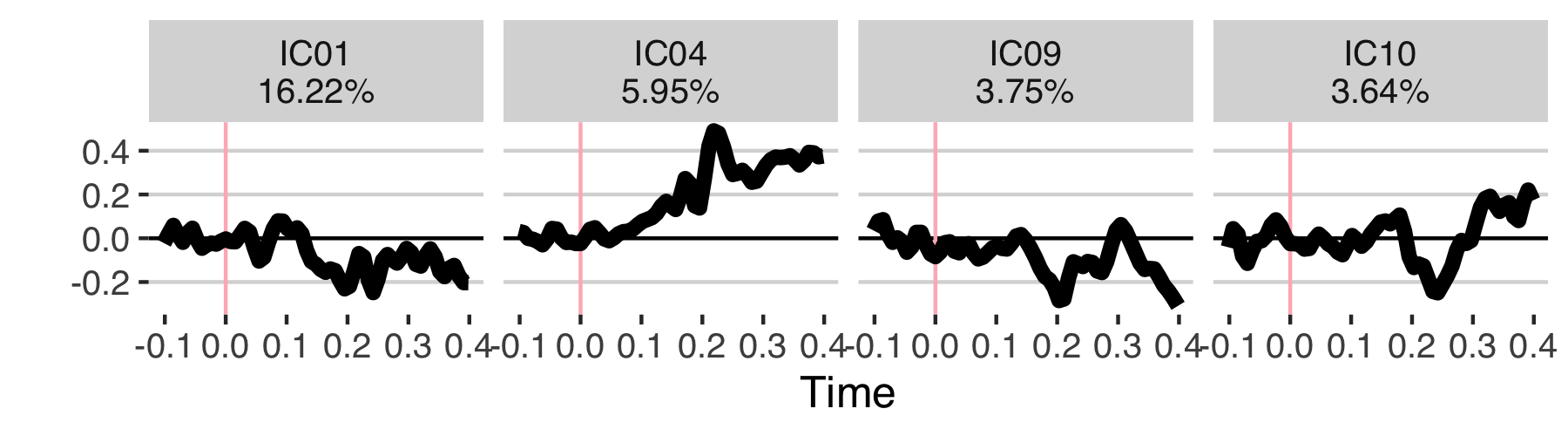}\\
\includegraphics[width=0.8\textwidth]{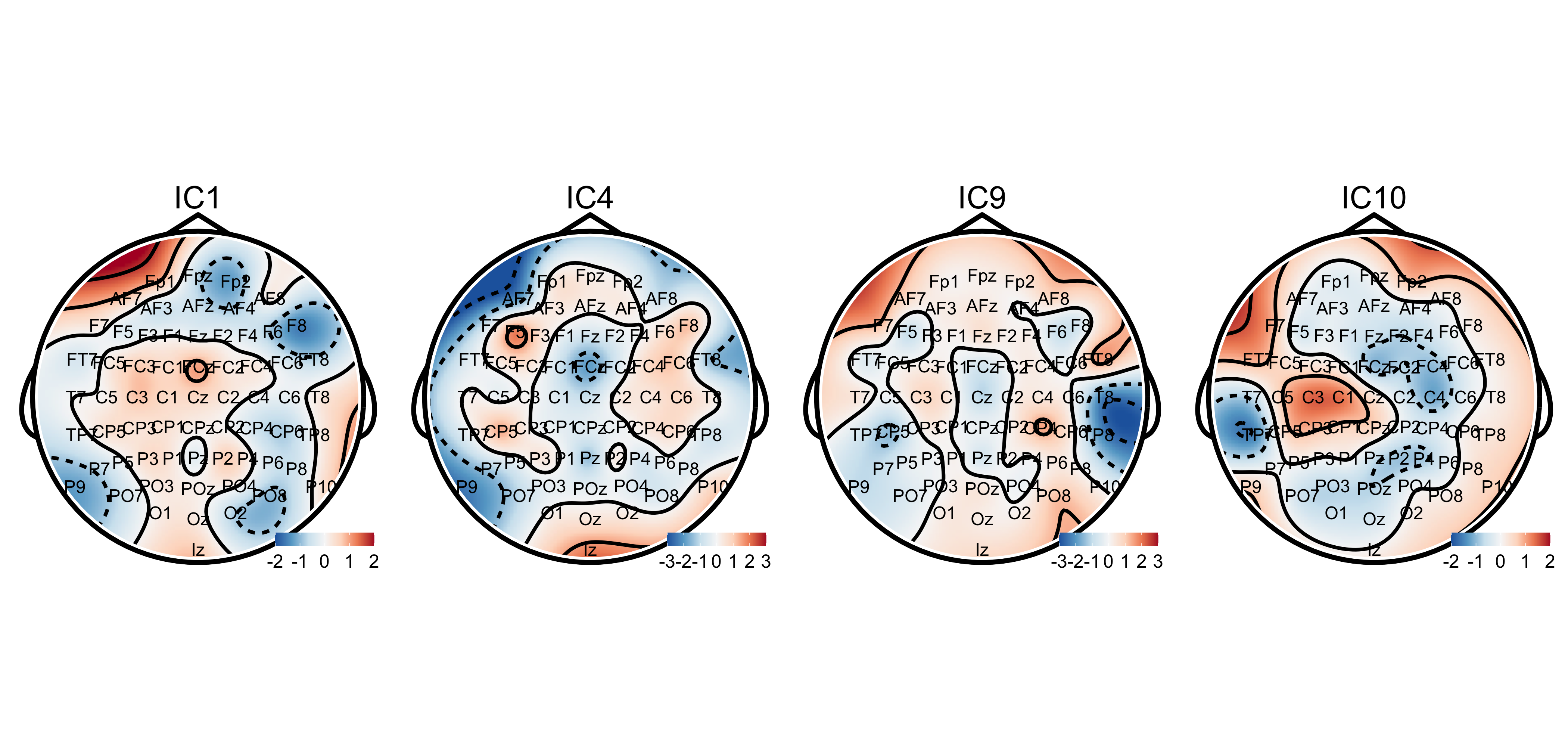}
\end{center}\caption{Topography maps of the four independent components identified as noise artifacts by cICA-LSP. ICs 1, 4, 9, and 10.}
\label{fig:eeglsp_artifact}
\end{figure}

\begin{figure}\begin{center}
\includegraphics[width=0.8\textwidth]{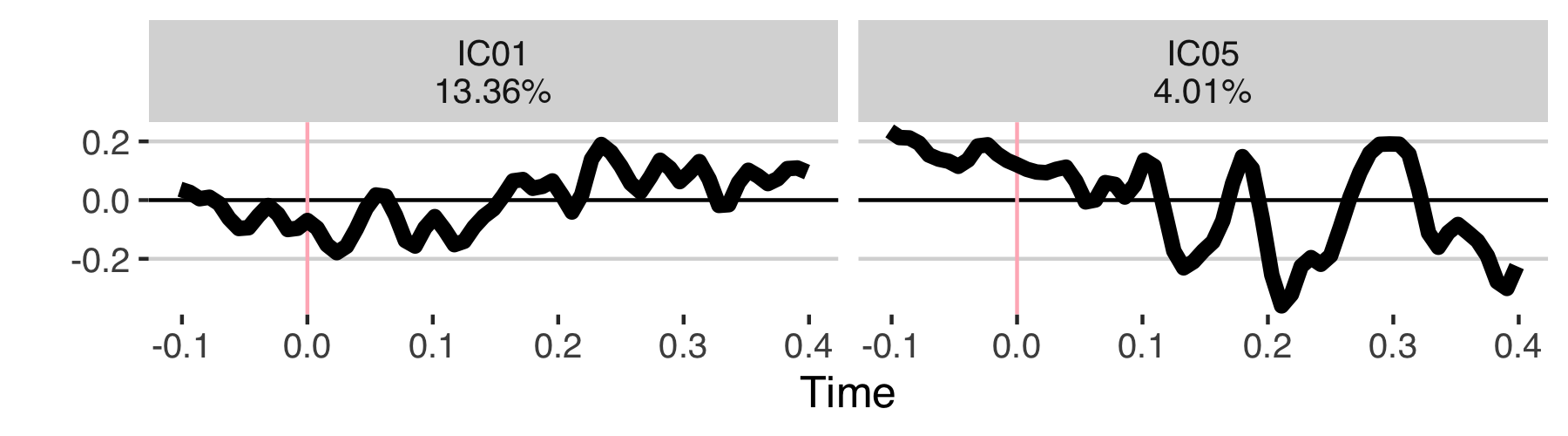}\\
\includegraphics[width=0.45\textwidth]{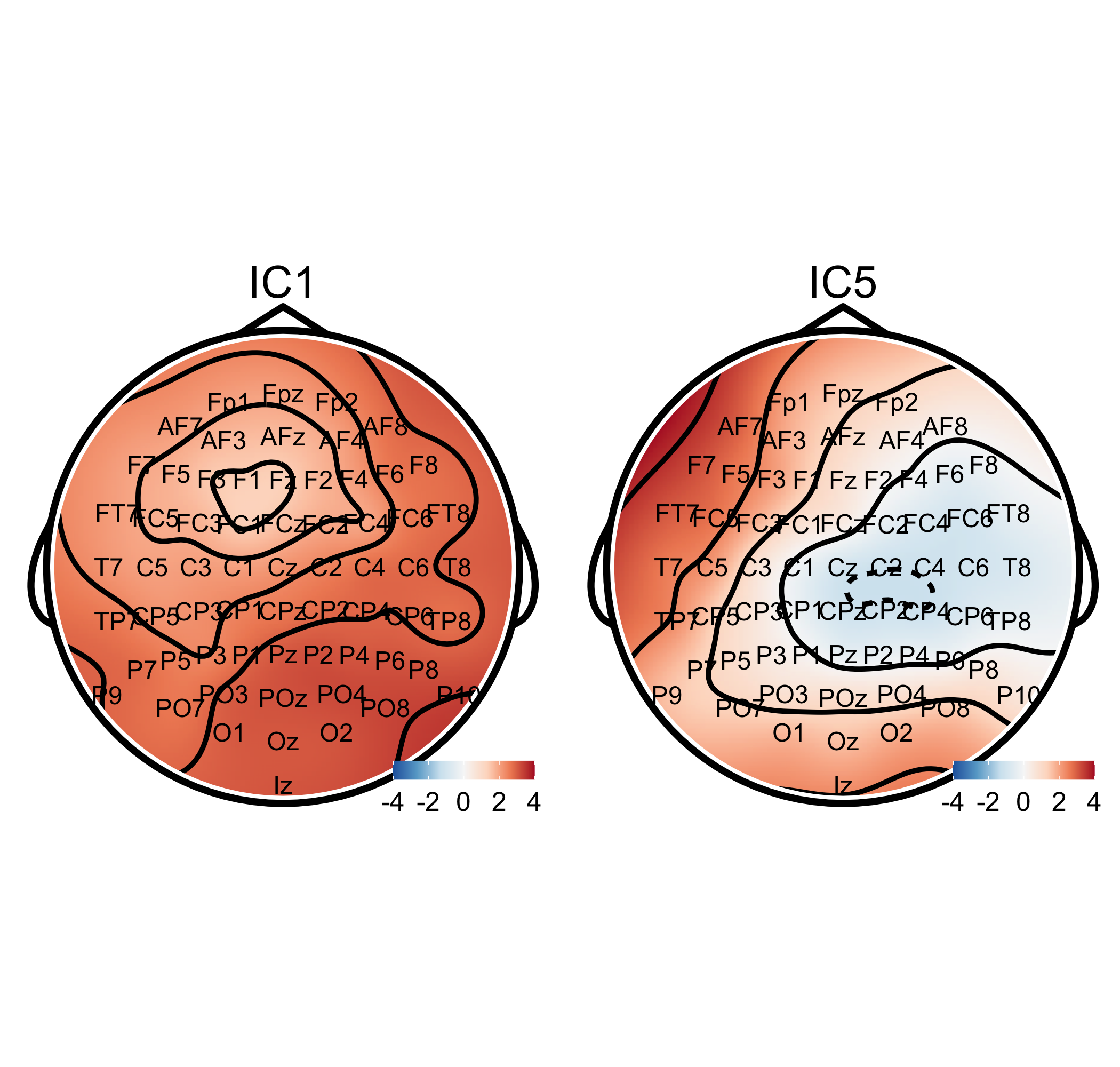}
\end{center}\caption{Topography maps of the two independent components identified as noise artifacts by SOBI.}
\label{fig:eegsobi_artifact}
\end{figure}

\begin{figure}\begin{center}
\includegraphics[width=0.99\textwidth]{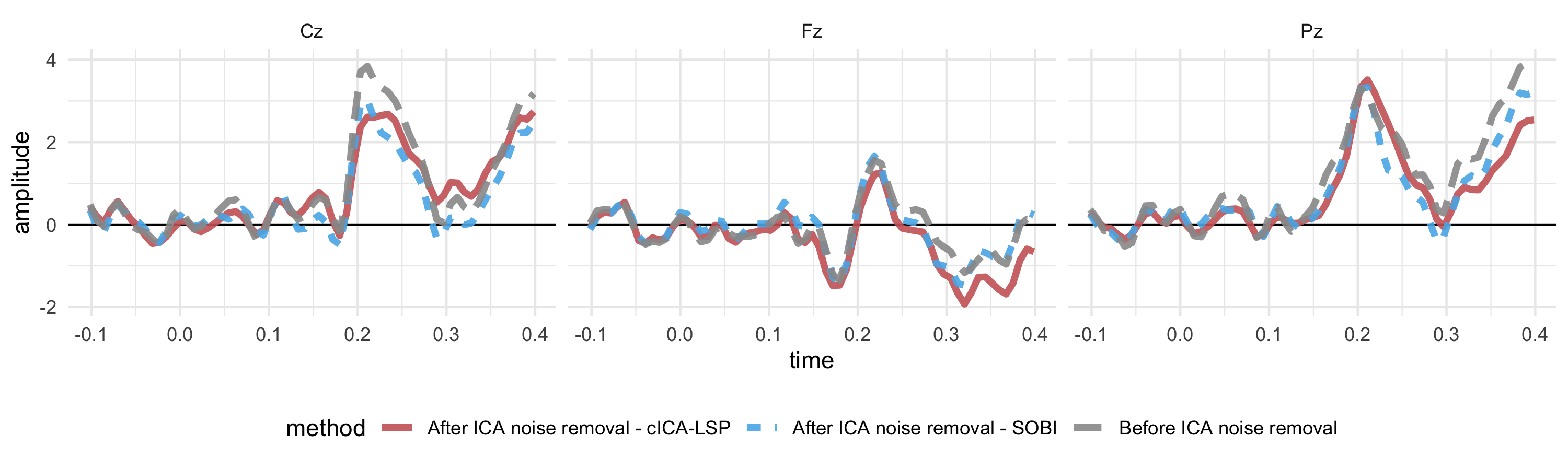}
\end{center}\caption{Average ERP before and after artifact removal at Cz, Fz and Pz.}
\label{fig:eeglsp_removed_3rois}
\end{figure}

The cICA-LSP identified IC 4 ($R^2$ = 0.182), 10 ($R^2$ = 0.136), 9 ($R^2$ = 0.135), and 1 ($R^2$ = 0.131) as noise artifacts. Those four components explained  29.56\% of the total variance of the EEG data, where each of them explained 5.95\%, 3.64\%, 3.75\%, and 16.22\% of the total variance, respectively. The SOBI identified IC 1 ($R^2$ = 0.183) and 5  ($R^2$ = 0.153) as the noise artifacts, which explained 17.37\% (13.36\% and 4.01\%, respectively) of the variance of the EEG data. Figures \ref{fig:eeglsp_artifact} and \ref{fig:eegsobi_artifact}  show the noise artifacts' average ERP and the corresponding topological maps. Figure \ref{fig:eeglsp_removed_3rois}  shows the average ERP before and after artifact removal at the three electrodes located in the midline (Cz, Fz, and Pz) for illustration. The noise artifacts were removed in Fz and Pz, while the signals were similar in Cz. The average ERP at 64 electrodes showed that ICA-based artifact identification removed the large variation toward the tailing signals (Figure \ref{fig:eeglsp_removed} in Appendix). After artifact removal, cICA and SOBI seem to have similar effects, while cICA identified more noise artifacts. 

\section{Consistency and Rate of Convergence}\label{lspec:sec:theory}
In this section, we report the theoretical results of cICA-LSP. We note that \cite{lee2020sampling} investigated the theoretical performance of the Whittle likelihood-based ICA when the source spectral density can be modeled with finite-dimensional parameters. Here we consider more general cases where the sources with mixed spectra, which finite-dimensional parameters may not effectively model. 

\subsection{Statement of the Results}
For each $j \in \{1,.\ldots,M\}$, consider a linear process $\{S_{j}(t)\}$ is a linear process,
$$ S_{j}(t) = \sum_{k=-\infty}^{\infty} a_{j,k} Z_{j}(t-k), \quad Z_{j}(t) \sim {i.i.d.}(0,\sigma_{j}^2),
$$
with spectral denstiy function
    $$ f_{j}(\lambda) = \frac{\sigma_{j}^2}{2 \pi}\left| \sum_{k=-\infty}^{\infty}
    a_{j,k} \exp (-ik \lambda) \right|^2, ~~~ -\pi \le \lambda \le \pi.
    $$
\begin{definition}
  A function $f$ on $[0,\pi]$ is said to satisfy {\it H\"{o}lder condition} with exponent $\gamma$ if there is a positive number $c$ such that $|f(x)-f(x_0)|\leq c\|x-x_0\|^{\gamma}$ for $x,x_0\in[0,\pi]$.
\end{definition}

\begin{definition}
  Let $m$ be a nonnegative integer and set $p=m+\gamma$. A function $f$ is said to be {\it p-smooth} if $f$ is $m$ times differentiable on $[0,\pi]$, and $f^{(m)}$ satisfies a  H\"{o}lder condition with exponent $\gamma$.
\end{definition}

The following condition ensures the spectral density functions of each source, $f_j$,  $j \in \{1,\ldots, M\}$, are $p$-smooth. 
\begin{condition}\label{lspecc2}
      For each $j$, $Z_{j}$ is white noise with finite moment and 
      $$\sum_{k} |a_{j,k}||k|^{p} < \infty, \quad p>\frac{1}{2}. $$
         \end{condition}
         
The following condition ensures that the logarithm of spectral density functions, $\varphi_j=\log f_j$, are bounded. This covers a wide variety of practical spectral densities, including AR or ARMA or the harmonic processes mixed with white noise as described in Examples 3.1 and 3.2.

    \begin{condition}\label{lspecc3}
      The power spectrum $f_{j}(\cdot)$, $j=1,\ldots, M$, are bounded away from zero.
          \end{condition}
	
The following condition on the specification of the number of knots $K$ is required for the desirable rate of convergence of the spline-based estimates of the power spectral densities.
  
	  \begin{condition}\label{lspecc4}
	    For each $j \in \{1,\ldots,M\}$, $K_j=O(K)$, where $K^{2} = \smallo{T^{1-\epsilon}}$ for some $\epsilon>0$.
          \end{condition}


To address permutation and scale ambiguity in ICA, several authors have used the idea of equivalence
class~\citep{chen2005cic,ilmonen2011semiparametrically,samworth2012independent,lee2020sampling}:
two matrices $\mb{W}$ and $\widetilde{\mb{W}}$ are equivalent, denoted as
$\mb{W} \sim \widetilde{\mb{W}}$, if $\widetilde{\mb{W}}=
\left(\mb{PD}\right)^{-1}\mb{W}$ for some permutation matrix $\mb{P}$ and a 
full-rank diagonal matrix $\mb{D}$. Given an equivalence class, the
unidentifiability issue can be resolved by choosing a representative of the class according to some
identifiability conditions proposed by \cite{chen2005cic}.  A full-rank matrix $\mb{W}$ satisfies the following conditions:
    \begin{enumerate}\label{identifiability}
      \item $\mb{W}_1 \prec \cdots \prec \mb{W}_{M}$ where $\mb{W}_{j}$
   is the $j$th row of $\mb{W}$ and we define $\prec$ as for
   $\forall \boldsymbol a, \boldsymbol b \in \mathbb{R}^{M}$
   iff there exists $k \in \{1,\ldots,M\}$ such that $k$th element of
   $\boldsymbol a$ is smaller than that of $\boldsymbol b$ and other
   element before $k$th are equal.
      \item $||\mb{W}_j||=1$ for $j \in \{1,\ldots,M\}$.
      \item $\max_{1\leq k \leq M} W_{jk} = \max_{1\leq k\leq M}
      |W_{jk}|$ for $j \in \{1,\ldots,M\}$.
    \end{enumerate}

    The following theorems show the consistency and the rate of convergence of the cICA-LSP estimate. 
\begin{theorem}\label{thm:consist}
      Let $\widehat{\mb{W}}$ be the cICA-LSP estimator and $\mb{W}_{0}$ be a true unmixing matrix. Under conditions \ref{lspecc2}-\ref{lspecc4} and the identifiability condition, $\widehat{\mb{W}}$ is consistent. 
        \begin{equation*}
	  \| \widehat{\mb{W}} - \mb{W}_{0} \|_{F} = o_{P}(1). 
	 \end{equation*}
       \end{theorem}
\begin{corollary}\label{thm:spec}
	Under the same condition of Theorem \ref{thm:consist}, the nuisance functions are also consistent. 
  \begin{equation*}
    \|\hat{\varphi}_{j}-\varphi \|=O_{P}\left(\sqrt{K/T}+K^{-p}\right), \quad j \in \{1,\ldots,M\}.
  \end{equation*}
\end{corollary}
	
\begin{theorem}\label{thm:conrate}
  Let $\widehat{\mb{W}}$ be a nonparametric-colorICA estimator and $\mb{W}_0$ be a true
    unmixing matrix. Then,
        \begin{equation*}
	  \|\hat{\mb{W}} - \mb{W}_{0} \|_{F} = O_{P}(T^{-1/2}). 
	\end{equation*}
     \end{theorem}

\subsection{Proofs}	  
	  The outline of results is as follows: First, we show the consistency and convergence rate of the Whittle likelihood estimate when the unmixing matrix is orthogonal. To ensure identifiability, we restrict the unmixing matric to the properly ordered space as \cite{chen2005cic}. Next, we extend the results to the full rank matrices satisfying the identifiability condition.
	  

\begin{lemma}\label{lem:r1:1}
  The expected value of the periodogram of $\mb{S}$ is $\text{diag}_j\{f_j(\rad)\}+\bigo{T^{-1}}$.
\end{lemma}
\begin{proof}
  The expectation of the periodogram is given as
  \begin{equation*}\begin{split}
    \E \PER(\rad,\mb{S})
    &=\frac{1}{2\pi T}\E \left(\sum_{t=1}^{T-1} \mb{S}(t)\exp(-i\rad t)\right)\left(\sum_{t=1}^{T-1} \mb{S}(t)\exp(-i\rad t)\right)^{*}\\
    &=\frac{1}{2\pi T}\sum_{t_1=0, t_2=0}^{T-1}\text{diag}_j \{\E S_j(t_1)S_j(t_2)\}\exp(-i\rad(t_1-t_2)).
  \end{split}\end{equation*}
For each $j$, we have
\begin{equation*}\begin{split}
  \frac{1}{2\pi T}\sum_{t_1=0, t_2=0}^{T-1}
  \{\E S_j(t_1)S_j(t_2)\}\exp(-i\rad(t_1-t_2))=\sum_{u=-T+1}^{T-1}\frac{T-|u|}{2\pi T}\gamma_j(u)\exp\{-i\rad u\}.
\end{split}\end{equation*}
By Condition \ref{lspecc2}, we have
\begin{multline}\label{eq:t1:bound}
\left|f_j(\rad)-\frac{1}{2\pi T}\sum_{t_1=0, t_2=0}^{T-1}
  \{\E S_j(t_1)S_j(t_2)\}\exp(-i\rad(t_1-t_2))\right|\\
\leq\frac{1}{2\pi}\sum_{|u|\geq T}\gamma_j(u)\exp\{-i\rad u\}
  + \frac{1}{2\pi}\sum_{|u|\leq T-1}\frac{|u|}{T}\gamma_j(u)\exp\{-i\rad u\}\\
\leq\frac{1}{2\pi}\sum_{|u|\geq T}\frac{|u|}{T}\gamma_j(u)\exp\{-i\rad u\}
  + \frac{1}{2\pi}\sum_{|u|\leq T-1}\frac{|u|}{T}\gamma_j(u)\exp\{-i\rad u\}=\bigo{T^{-1}},
\end{multline}
and the desired results hold.
\end{proof}

Denote the true unmixing matrix as $\mb{W}_0$. From Lemma \ref{lem:r1:1}, the expected value of the Whittle likelihood is give by
\begin{equation} \label{eq:t1:expectedWL}
      \begin{split}
	\E \mc{L}_{T}(\mb{W},\boldbeta) 
      &= -\frac{1}{[T/2]}  \sum_{j=1}^M \sum_{k=1}^{[T/2]} 
      \left \{ \frac{\mb{e}_{j}^{\top} \mb{U} \mb{F}_{S}(\rad_k) \mb{U}^{\top} \mb{e}_{j}}{\exp\{g_j(\rad_k;\boldbeta_j)\}} +  g_j(\rad_k;\boldbeta_j) \right \}\\
      & \quad + T\ln{|\det{\mb{W}}|} + \bigo{T^{-1}},
    \end{split}
    \end{equation}
    where $\mb{U} = \mb{WW}_{0}^{-1}$. Since we restrict $\mb{W}$ to orthogonal matrices, we omit the term $\ln|\text{det}\mb{W}|$ rest of this paper. Let the asymptotic expected Whittle log-likelihood written as
    \begin{equation} \label{eq:r1:approxex}
      \begin{split}
       \Lambda_T(\mb{W},\boldbeta)
      &= -\frac{1}{[T/2]} \sum_{j=1}^M \sum_{k=1}^{[T/2]} 
      \left \{ \frac{\mb{e}_{j}^{\top} \mb{U} \mb{F}_{S}(\rad_k) \mb{U}^{\top} \mb{e}_{j}}{\exp\{g_j(\rad_k;\boldbeta_j)\}} +  g_j(\rad_k;\boldbeta_j) \right \},
     \end{split}
    \end{equation}
    By Jensen's inequality and orthogonality of $\mb{W}_0$ and $\mb{W}$, $\Lambda_T(\mb{W},\boldbeta)$ is bounded by $ \Lambda_T(\mb{W}_0, \boldvarphi)$:
   \begin{equation} 
      \begin{split}
       \Lambda_T(\mb{W},\boldbeta)
      & \leq -\frac{1}{[T/2]} \sum_{j=1}^M \sum_{k=1}^{[T/2]}  \left \{ 1 + \sum_{l=1}^M\log{U_{jl}^2 f_l(\rad_k)} \right \}\\
     & \leq -\frac{1}{[T/2]} \sum_{k=1}^{[T/2]} \sum_{l=1}^M  \left \{ 1 + \sum_{j=1}^MU_{jl}^2\log{f_l(\rad_k)} \right \}= \Lambda_T(\mb{W}_0, \boldvarphi).
  \end{split}
    \end{equation}

\subsubsection{Bias}

For a given orthogonal matrix $\mb{W}$, write $\varphi_j^W = \log(\sum_{m=1}^M U_{jm}^2 f_m)$, where $\mb{U}=\mb{W}\mb{W}_0^{-1}$. 
\begin{lemma}\label{lemma:2}
      Let $\epsilon$ be a positive constant. There exist postive constants $c_1$ and $c_2$ such that
      $$\|\varphi_j^W-\varphi_j\|^2\leq c_1 \|\mb{W}-\mb{W}_0\|_F^2,~\text{and~}
      \|\varphi_j^W-\varphi_j\|_{\infty}\leq c_2 \|\mb{W}-\mb{W}_0\|_F^2,$$
      for $\epsilon$ small enough and any orthogonal matrix $\mb{W}$ satisfying $\|\mb{W}-\mb{W}_0\|_F\leq \epsilon$.
    \end{lemma}
    \begin{proof}

    For a give $\epsilon>0$, let $\mb{W}$ be close to $\mb{W}_0$ such that $\|\mb{W}-\mb{W}_0\|_F<\epsilon$. Let the difference matrix $\bolddelta$ satisfy $\mb{U}=\mb{WW}_0^{-1}=\mb{I}+\bolddelta$. This leads $\mb{W}=\mb{W}_0+\bolddelta\mb{W}_0$. Then, we have $M-\epsilon\leq \tr{\mb{U}}\leq M$ and  $-\epsilon^2/2\leq \tr{\bolddelta}\leq 0$. Since $\mb{U}$ is orthogonal, we have $-2\tr{\bolddelta}=\sum_{j,m}\delta_{jm}^2$. Thus, the sum of squares of the difference matrix is bounded as $0\leq \sum_{j,m}\delta_{jm}^2\leq \epsilon^2$. Note that from the orthogonality condition, all the diagonal elements of $\bolddelta$ have negative value.\footnote{Assume that there is at least one $j$ such that $\delta_{jj}>0$. Since $\mb{U}$ is orthogonal, $\|\mb{U}_j\|^2=1+2\delta_{jj}+\sum_{m=1}^M \delta_{jm}^2=1$. The equality holds only if $\delta_{jj}=0$, which contradicts the assumption.} 

For each $j=1,\ldots,M$,
\begin{equation}\label{eq:bound}\begin{split}
      \|\varphi_j^W-\varphi_{j}\|^2
      &=\int\left(\log\left(\frac{\sum_{m=1}^MU_{jm}^2f_m(\rad)}{f_j(\rad)}\right)\right)^2 d\rad\\
      &=\int\left(\log\left(1+2\delta_{jj}+\frac{\sum_{m=1}^M \delta_{jm}^2f_m(\rad)}{f_j(\rad)}\right)\right)^2d\rad
        \end{split}
    \end{equation} 
    Then the righthand side of \refp{eq:bound} is bounded by
\begin{equation}\begin{split}
  \refp{eq:bound}
  &\leq\int \left| 2\delta_{jj}+\sum_{m=1}\delta_{jm}^2\frac{f_m(\rad)}{f_j(\rad)}\right| d\rad
  \leq b_1\int 2|\delta_{jj}|+\left(\frac{\sum_{m=1}^M \delta_{jm}^2f_m(\rad)}{f_j(\rad)}\right)^2d\rad \\
  & \leq b_1\left( -2\sum_{j=1}^M \delta_{jj} + b_2  \sum_{j=1^M}\delta_{jm}^2\right)
  \leq b_1(\epsilon^2 + b_2\epsilon^2)=b_1(1+b_2)\|\mb{W}-\mb{W}_0\|_F^2,
    \end{split}
    \end{equation} 
    where $b_1$ is a positive constant and $b_2=\frac{1}{2\pi}\max_{m,j}\sup_{\rad} \frac{f_m(\rad)}{f_j(\rad)}$. 

    Similarly, 
    \begin{equation}\begin{split}
      \|\varphi_j^W-\varphi_{j}\|_{\infty}
      &=\int\sup_r\left|\ln\left(1+2\delta_{jj}+\frac{\sum_{m=1}^M \delta_{jm}^2f_m(\rad)}{f_j(\rad)}\right)\right|d\rad\\
      &\leq b_3\int 2|\delta_{jj}|+\sum_{m=1}^M \delta_{jm}^2\sup_r \frac{f_m(\rad)}{f_j(\rad)}d\rad\\
      &\leq b_3 \left(-2\sum_{j=1}^M \delta_{jj}\right) + b_4  \sum_{j=1}^M \delta_{jm}^2 \leq (b_3+b_4)\|\mb{W}-\mb{W}_0\|_F^2.
    \end{split}
    \end{equation} 
    Thus, the desired results hold.
  \end{proof}

  Write $\hat{\boldbeta}^W =\arg\max_{(\boldbeta: g(\cdot;\boldbeta_j)\in s, j=1,\ldots,M)} L_T(\mb{W},\boldbeta)$ and $\hat{\varphi}_j^W=g(\cdot;\hat{\boldbeta}_j^W)$. By the Theorem of \cite{kooperberg1995rcl}, for any given orthogonal matrix $\mb{W}$, $\hat{\varphi}^W$ is a consistent estimate of $\varphi^W$ such that
    \begin{equation*}
      \|\hat{\varphi}_j^W-\varphi_j^W\| = O_p\left(\sqrt{\frac{K}{T}}+K^{-p}\right), \quad j=1,\ldots,M.
    \end{equation*}
  From Lemma \ref{lemma:2}, if $\mb{W}$ is close to $\mb{W}_0$, for a positive constant $c$,
    \begin{equation*}\begin{split}
      \|\hat{\varphi}_j^W-\varphi_j\|
      &\leq \underbrace{\|\hat{\varphi}_{j}^W-\varphi_{j}^W\|}
      +\underbrace{\|\varphi_{j}^W-\varphi_{j}\|}\\
      & \leq O_p\left(\sqrt{\frac{K}{T}}+K^{-p}\right) + c\|\mb{W}-\mb{W}_0\|_F^2
    \end{split}\end{equation*}
    This indicates that the bias depends on that of $\widehat{\mb{W}}$. 

Let $\tau_T$, $T\geq 1$, be positive numbers such that $K \tau_T^2=O(1)$ and $\frac{1}{T} = o\left(\tau_T^2\right)$.

\begin{lemma}\label{lemma:rate}
	Let $b_1$, $b_2$ and $b_3$ be positive constants. There is a positive 
	constant $b_4$ such that, for $T$ sufficiently large,
	\begin{equation*}
	  \Pr \left(\left|L_T(\mb{W},g)-L_T(\mb{W}^*,g^*)
	  -\Lambda_T(\mb{W},g)+\Lambda_T(\mb{W}^*,g^*) \right|\geq b_1 \tau_T\right)\leq 2 \exp\left(-b_4 T\tau_T^2\right)
	\end{equation*}
	for any $\mb{W}$ satisfying $\|\mb{W}-\mb{W}^*\|_F\leq b_2 \tau_T$ and $g_j$ satisfying $\|g_j-g_j^*\|_F\leq b_3 \tau_T$.
\end{lemma}
\begin{proof}
  By Hoeffding's inequality, for any positive constant $t>0$, there is a positive constant $c$ such that 
  \begin{equation*}\begin{split}
    &\Pr \left(\left|L_T(\mb{W},g)-L_T(\mb{W}^*,g^*)
	  -\Lambda_T(\mb{W},g)+\Lambda_T(\mb{W}^*,g^*) \right|\geq b_1 \tau_T\right)\\
	  & =\Pr \left(\exp\left\{t\left|L_T(\mb{W},g)-L_T(\mb{W}^*,g^*)
	  -\Lambda_T(\mb{W},g)+\Lambda_T(\mb{W}^*,g^*)\right|\right\} \geq \exp\{tb_1\tau_T\} \right)\\
	  & \leq 2 \left\{1 + t^2 \frac{c}{2T}\right\}\exp\{-tb_1\tau_t\}\\
	  & \leq 2 \exp\left\{t^2 \frac{c}{2T}-tb_1\tau_t\right\}.
	\end{split}\end{equation*}
	If we set $t= \frac{Tb_1\tau_T}{c}$, then the desired results hold.
\end{proof}

Write $g_j^{(l)}(\cdot)=g\left(\cdot;\boldbeta_j^{(l)}\right)$, for $j=1,\ldots,M$ and $l=1,2$.
\begin{lemma}\label{lemma:approx}
	Let $\epsilon$, $b_1$, $b_3$ and $b_2$ be positive constants. Then, except on an event with probability tending to zero as $T\rightarrow\infty$, 
	\begin{equation*}
	  \left|L_T\left(\mb{W}_1,g^{(1)}\right)-L_T\left(\mb{W}_2,g^{(2)}\right)\right|\leq\epsilon\tau_T
	\end{equation*}
	for all $\|\mb{W}_1-\mb{W}_2\|_F\leq b_1\tau_T$, $\|g_j^{(1)}-g_j^{(2)}\|_{\infty} \leq b_2\tau_T$, $\|g_j^{(1)}\|_{\infty}\leq b_3$ and $\|g_j^{(2)}\|_{\infty}\leq b_3$.
      \end{lemma}
\begin{proof}
  Write $\mb{G}_l(\rad_k)=\text{diag}_{j=1,\ldots,M}\{g_j^{(l)}(\rad_k)\}$, $l=1,2$.
   \begin{equation*}\begin{split}
    &\left|L_T\left(\mb{W}_1,g^{(1)}\right)-L_T\left(\mb{W}_2,g^{(2)}\right)\right|\\
     &\leq \frac{1}{2T}\sum_{j,k}\left|\tr{\PER(\rad_k;\mb{X})
     \left\{\mb{W}_1^{\top}\mb{G}_1^{-1}(\rad_k)\mb{W}_1-\mb{W}_2^{\top}\mb{G}_2^{-1}(\rad_k)\mb{W}_2\right\}}\right|
     + \frac{1}{2T}\sum_{j,k}\left| g_j^{(1)}(\rad_k)-g_j^{(2)}(\rad_k)\right|\\
     &\leq \frac{1}{2T}\sum_{j,k}\left|\tr{\PER(\rad_k;\mb{X})
    \left\{\mb{W}_1^{\top}\mb{G}_1^{-1}(\rad_k)\mb{W}_1-\mb{W}_1^{\top}\mb{G}_2^{-1}(\rad_k)\mb{W}_1\right\}}\right|\\
    &\quad+ \frac{1}{2T}\sum_{j,k}\left|\tr{\PER(\rad_k;\mb{X})
    \left\{\mb{W}_1^{\top}\mb{G}_2^{-1}(\rad_k)\mb{W}_1-\mb{W}_1^{\top}\mb{G}_2^{-1}(\rad_k)\mb{W}_2\right\}}\right|\\
    &\quad+ \frac{1}{2T}\sum_{j,k}\left|\tr{\PER(\rad_k;\mb{X})
    \left\{\mb{W}_1^{\top}\mb{G}_2^{-1}(\rad_k)\mb{W}_2-\mb{W}_2^{\top}\mb{G}_2^{-1}(\rad_k)\mb{W}_2\right\}}\right|
    + M b_2\tau_T\\
    &= O_p(1) (b_2 + 2b_1b_3)\tau_T
   \end{split}\end{equation*}
\end{proof}

\begin{proof}[Proof of Theorem 1]

Write $\boldtheta=(\text{vec}\left(\mb{W}\right)^{\top},\boldbeta^{\top})^{\top}$. 
From Lemmas \ref{lemma:rate} and \ref{lemma:approx}, except on an event whose probability tends to zero as $T\rightarrow \infty$, $L_T(\mb{W}^{(1)},g^{(1)})< L_T(\mb{W}^*,g^*)$ for $\|\mb{W}^{(1)}-\mb{W}^*\|_F=b_1\tau_T$ and $\|g_j^{(1)}-g^*_j\| =b_2\tau_T$. 
For fixed constants $b_1$ and $b_2$, denote an open ball at $\boldtheta$ with radius $\tau_T$ as $N(\boldtheta; \tau_T)=\{\boldtheta^{(1)}: \|\mb{W}^{(1)}-\mb{W}\|_F \leq b_1\tau_T,~\|g_j^{(1)}-g_j\|\leq b_2\tau_T\}$. Then we have 
\begin{equation}\label{eq:510}
  \lim_{T \rightarrow \infty}\Pr\left\{\sup_{\boldtheta \in N(\boldtheta_1;\tau_T)}L_T(\mb{W},g) - L_T(\mb{W}^*,g^*) <0 \right\}=1.
\end{equation}

Let $B=\{(\mb{W},\boldbeta):\|\mb{W}-\mb{W}_0\|_F\geq b_1\tau_T,~\|g_j-g^*_j\| \geq b_2\tau_T\}$. Since $B$ is compact, there exists a collection of finite number of open coverings: $\{N(\boldtheta_p; \delta_p): p=1,\ldots,P\}$.

By \refp{eq:510} and finite number $P$, we have 
\begin{equation*}
  \lim_{T \rightarrow \infty}\Pr\left\{\sup_{ \boldtheta\in \cup_{p=1}^P N(\boldtheta_p; \delta_p)}L_T(\mb{W},g) - L_T(\mb{W}^*,g^*) <0 \right\}=1,
\end{equation*}
thus,
\begin{equation*}
  \lim_{T \rightarrow \infty}\Pr\left\{\sup_{\boldtheta }L_T(\mb{W},g) =\sup_{ \boldtheta \in N(\theta^*;\tau_T)}L_T(\mb{W},g)   \right\}=1,
\end{equation*}
Therefore, $\|\widehat{\mb{W}}-\mb{W}^*\|=o_P(\tau_T)=o_P(1)$ and $\|\hat{g}_j-g_j^*\|=o_P(\tau_T)$. 
From \refp{eq:r1:approxex}, the definition of $\mb{W}^*$ and $g^*$ 
and Lemma 2 from \citep{kooperberg1995rcl},
\begin{equation}\label{eq:maximum}\begin{split}
  0& \leq\Lambda_T(\mb{W}^*,g^*)-\Lambda_T(\mb{W}_0,g^{W_0})
\leq \Lambda_T(\mb{W}_0,\varphi)-\Lambda_T(\mb{W}_0,g^{W_0})\\
& \leq c \sum_{j=1}^M\left\|\varphi_j-g^{W_0}\right\|= O(K^{-2p}).
 \end{split}\end{equation}

Let $\dot{\Lambda}_T(\boldtheta)=\frac{\partial\Lambda_T(\boldtheta)}{\partial \boldtheta}$. 
We can write 
$\int_0^1 \frac{\partial}{\partial u}\Lambda_T(\boldtheta^* + u(\boldtheta_0-\boldtheta^*))du =\Lambda_T(\mb{W}^*,g^*)-\Lambda_T(\mb{W}_0,g^{W_0})$. This can further be written as $\mb{D}\left(\boldtheta^*-\boldtheta_0\right)=\Lambda_T(\mb{W}^*,g^*)-\Lambda_T(\mb{W}_0,g^{W_0})$, where $\mb{D}$ is given by $\mb{D}=\int_0^1  \dot{\Lambda}_T(\boldtheta^* + u(\boldtheta_0-\boldtheta^*))du$. Since $\mb{D}$ is bounded, we have 
$\|\boldtheta^*-\boldtheta_0\|= O(K^{-2p})$. 

Thus, $\|\widehat{\mb{W}}-\mb{W}_0\|=o_P(1)$ and $\|\hat{g}_j-\varphi_j\|=o_P(1)$.
\end{proof}

\subsection{Variance}

Let $\dot{L}_T(\boldtheta)$ denote the score at $\boldtheta$ of $(M^2+MK)$-dimensional vector with entries $\frac{\partial L_T(\boldtheta)}{\partial \theta_j}$:
\begin{equation}\begin{split}
  \frac{\partial L(\boldtheta)}{\partial\mb{W}_j}
  &=-2\frac{1}{[T/2]}\sum_k \PER(\rad_k;\mb{X})\mb{W}\mb{e}_j\exp\{-g_j(\rad_k;\boldbeta_j)\}\\
  \frac{\partial L(\boldtheta)}{\partial\boldbeta_j}
  &=\frac{1}{[T/2]}\sum_k \text{tr}\left(\mb{W}\PER(\rad_k;\mb{X})\mb{W}^{\top}\right)\exp\{-g_j(\rad_k;\boldbeta_j)\}\mb{B}_j(\rad_k) -  \frac{1}{[T/2]}\sum_k \mb{B}_j(\rad_k).
\end{split}\end{equation}
Let $\mb{H}_T(\boldtheta)$ denote the Hessian at $\boldtheta$; that is, the $(M^2+\sum_j K_j)\times(M^2+\sum_j K_j)$ matrix with entries $\frac{\partial^2 L_T(\boldtheta)}{\partial \theta_j\partial \theta_k}$:
\begin{equation}\begin{split}
\frac{\partial^2 L(\boldtheta)}{\partial\mb{W}_j\partial\mb{W}_k}
  &=-\frac{2}{[T/2]}\sum_k \delta_{j-k} \PER(\rad_k;\mb{X}) \exp\{-g_j(\rad_k;\boldbeta_j)\}\\
  \frac{\partial^2 L(\boldtheta)}{\partial\mb{W}_j\partial\boldbeta_k}
  &=\frac{2}{[T/2]}\sum_k \delta_{j-k} \left(\PER(\rad_k;\mb{X})\mb{W}\mb{e}_j\right)\exp\{-g_j(\rad_k;\boldbeta_j)\}\mb{B}_j(\rad_k)\\
    \frac{\partial^2 L(\boldtheta)}{\partial\boldbeta_j\partial\boldbeta_k}
    &=\frac{1}{[T/2]}\sum_k \delta_{j-k} \text{tr}\left(\mb{W}\PER(\rad_k;\mb{X})\mb{W}^{\top}\right)\exp\{-g_j(\rad_k;\boldbeta_j)\}\mb{B}_j(\rad_k) \mb{B}_j(\rad_k)^{\top}. 
\end{split}\end{equation}
Then, we have
\begin{equation}\label{eq:variance:score} \int_0^1\frac{d}{du}\dot{L}_T\left(\boldtheta^*+u(\hat{\boldtheta}-\boldtheta^*)\right)du=\dot{L}_T(\hat{\boldtheta})-\dot{L}_T(\boldtheta^*). \end{equation}
  Define a $(M^2+MK)\times(M^2+MK)$ matrix $\mb{D}=\int_0^1\mb{H}_T\left(\boldtheta^*+u(\hat{\boldtheta}-\boldtheta^*)\right)du$. Then \refp{eq:variance:score} can be written as $\mb{D}(\hat{\boldtheta}-\boldtheta^*)=\dot{L}_T(\hat{\boldtheta})-\dot{L}_T(\boldtheta^*)$. Since $(\hat{\boldtheta}-\boldtheta^*)^{\top}\dot{L}_T(\hat{\boldtheta})=0$ and $(\hat{\boldtheta}-\boldtheta^*)^{\top}\E\dot{L}_T(\hat{\boldtheta})=0$, we have
  \begin{equation}
  (\hat{\boldtheta}-\boldtheta^*)^{\top}\mb{D}(\hat{\boldtheta}-\boldtheta^*)^{\top}=-(\hat{\boldtheta}-\boldtheta^*)^{\top}\left(\dot{L}_T(\boldtheta^*)-\E\dot{L}_T(\boldtheta^*)\right).
  \end{equation}

  \begin{lemma}\label{lemma:variance1}
    Under Condition 1, we have
    \[ |\dot{L}_T(\boldtheta^*)-\E\dot{L}_T(\boldtheta^*)|^2=O_P\left(\frac{1}{T}\right).\]
  \end{lemma}

  \begin{lemma}\label{lemma:variance2}
    There exists a positive constant $c$ such that 
    \[(\hat{\boldtheta}-\boldtheta^*)^{\top}\mb{D}(\hat{\boldtheta}-\boldtheta^*)^{\top}
    \leq \frac{cK^{-1}}{T}\|\hat{\boldtheta}-\boldtheta^*\|^2, \]
    except on an event whose probability ends to zero as $T\rightarrow\infty$.
  \end{lemma}

  \begin{proof}
Write $\mb{a}=\hat{\boldtheta}-\boldtheta^*$, $\mb{a}^W=\left(\vect(\widehat{\mb{W}})-\vect(\mb{W}^*)\right)$, $\mb{a}_j^{\beta}=\left(\hat{\boldbeta}_j-\boldbeta_j^*\right)$. For any $\boldtheta$, 
\begin{eqnarray}
  \mb{a}^{\top}\mb{H}_T(\boldtheta)\mb{a}
  =-\sum_{j=1}^M \frac{2}{[T/2]} {\mb{a}_j^W}^{\top}\sum_k \text{Re}\PER(\rad_k;\mb{X})
  \exp\{-\boldbeta_j^{\top}\mb{B}_j(\rad_k)\} \mb{a}_j^W \label{eq:variance:1}\\
   + 2\sum_{j=1}^M \frac{2}{[T/2]} {\mb{a}_j^W}^{\top}\sum_k \text{Re}\PER(\rad_k;\mb{X})\mb{W}\mb{e}_j^{\top}
  \exp\{-\boldbeta_j^{\top}\mb{B}_j(\rad_k)\} \mb{B}_j^{\top}(\rad_k)\mb{a}_j^{\beta} \label{eq:variance:2}\\
   - \sum_{j=1}^M \frac{1}{[T/2]} {\mb{a}_j^{\beta}}^{\top}\text{tr}\left(\mb{W}\PER(\rad_k;\mb{X})\mb{W}^{\top}\right)\exp\{-g_j(\rad_k;\boldbeta_j)\}\mb{B}_j(\rad_k) \mb{B}_j(\rad_k)^{\top} \mb{a}_j^{\beta}  \label{eq:variance:3}
\end{eqnarray}

From the asymptotic distribution of the periodogram, we have 
\begin{equation}\label{eq:variance:aha1}
  \frac{2}{[T/2]}\sum_k \text{Re}\PER(\rad_k;\mb{X})\exp\{-\boldbeta_j^{\top}\mb{B}_j(\rad_k)\}=O_P(T^{-1/2}).
\end{equation}
  
  Then there is a positive constant $c_1$, such that 
\[\sum_{j=1}^M \frac{2}{[T/2]} {\mb{a}_j^W}^{\top}\sum_k \text{Re}\PER(\rad_k;\mb{X})
\exp\{-\boldbeta_j^{\top}\mb{B}_j(\rad_k)\} \mb{a}_j^W \geq c_1 \|\mb{a}^W\|^2\geq c_1 K^{-1}\|\mb{a}^W\|^2.\] 
From the Lemma 6 in \cite{kooperberg1995rcl}, there is a positive constant $c_2$ such that 
\[\sum_{j=1}^M \frac{1}{[T/2]} \text{tr}\left(\mb{W}\PER(\rad_k;\mb{X})\mb{W}^{\top}\right)\exp\{-g_j(\rad_k;\boldbeta_j)\} \left|\left(\hat{\boldbeta}_j-\boldbeta_j^*\right)^{\top}\mb{B}_j(\rad_k)\right|^2 \geq c_2 K^{-1}\|\hat{\boldbeta}_j-\boldbeta_j^*\|^2.\]
Also, there exists a positice constant $c_3$ such that
\begin{equation*}\begin{split}
  |\refp{eq:variance:2}|
  &\leq \sqrt{2}\sum_{j=1}^M \left\|{\mb{a}_j^W}\right\|\left\|  \frac{2}{[T/2]} \sum_k \text{Re}\PER(\rad_k;\mb{X})\mb{W}\mb{e}_j^{\top}
  \exp\{-\boldbeta_j^{\top}\mb{B}_j(\rad_k)\} \mb{B}_j^{\top}(\rad_k) \right\|_F\left\|\mb{a}_j^{\beta}\right\|\\
  &\leq c_3 \left\|\mb{a}\right\|^2.
\end{split}\end{equation*}

  Thus, there exists a positive constant $c_4$ such that 
  $\mb{a}^{\top}\mb{H}_T(\boldtheta)\mb{a}\leq -c_4 K^{-1}\|\mb{a}\|^2$. 
\end{proof}

Since \[ \left|(\hat{\boldtheta}-\boldtheta^*)^{\top}\left(\dot{L}_T(\boldtheta^*)-\E\dot{L}_T(\boldtheta^*)\right)\right|
\leq \left|\hat{\boldtheta}-\boldtheta^*\right| \left|\dot{L}_T(\boldtheta^*)-\E\dot{L}_T(\boldtheta^*)\right|,  \]
and from Lemmas \ref{lemma:variance1}--\ref{lemma:variance2}, we have 
\begin{equation}
  \|\hat{\boldtheta}-\boldtheta^*\|^2=O_P(K^2/T),
\end{equation}

\section{Conclusion}
This paper proposed a new independent component analysis for the auto-correlated sources with mixed spectra. We derived the Whittle likelihood function to handle autocorrelation. We also employed nonparametric spectral density estimation to model the mixed spectra accurately. Numerical experiments showed superior performance of recovering sources compared to existing ICA methods and supported the theoretical findings. Furthermore, EEG data application demonstrated its effectiveness in applications.

\section*{Appendix}

\subsection*{Simulation Results}

\begin{figure}
\centering
\includegraphics[width=0.8\textwidth, height=0.4\textheight]{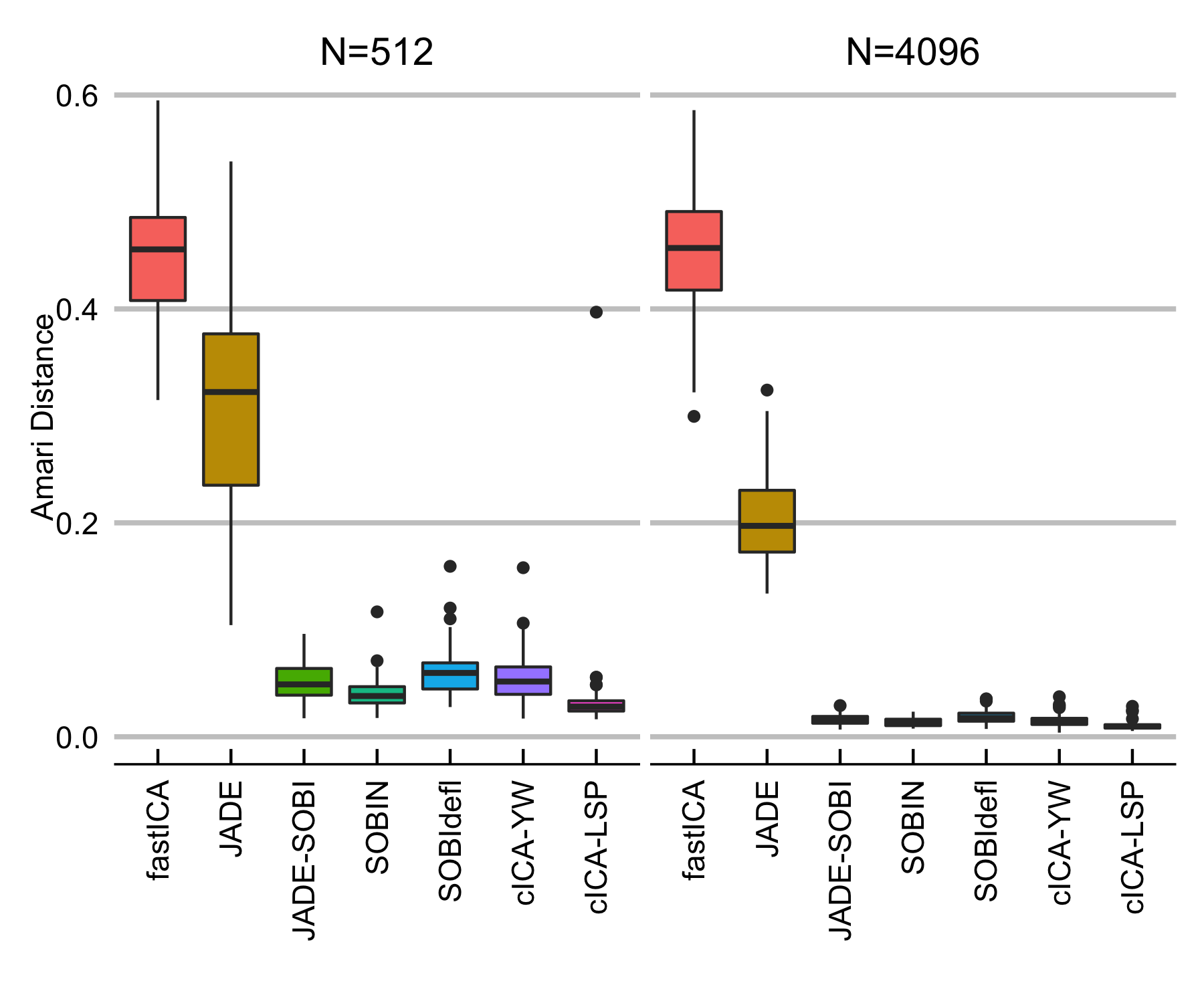}
\caption[Simulation Study I: Performance comparison for sources with mixed spectra of atoms at Fourier frequency]
{Simulation Study I: Full performance comparisons including JADE and fastICA in addition to those presented in Figure \ref{fig:sim:amari}.}\label{fig:sim:amarifull}
\end{figure}

Figure \ref{fig:sim:amarifull} includes the performance of JADE and fastICA, in addition to those shown in Figure \ref{fig:sim:amari}. JADE and fastICA performed much worse than the ICA methods account for autocorrelation.

\subsection*{EEG Data Analysis Results}
\begin{figure}\begin{center}
\includegraphics[width=\textwidth]{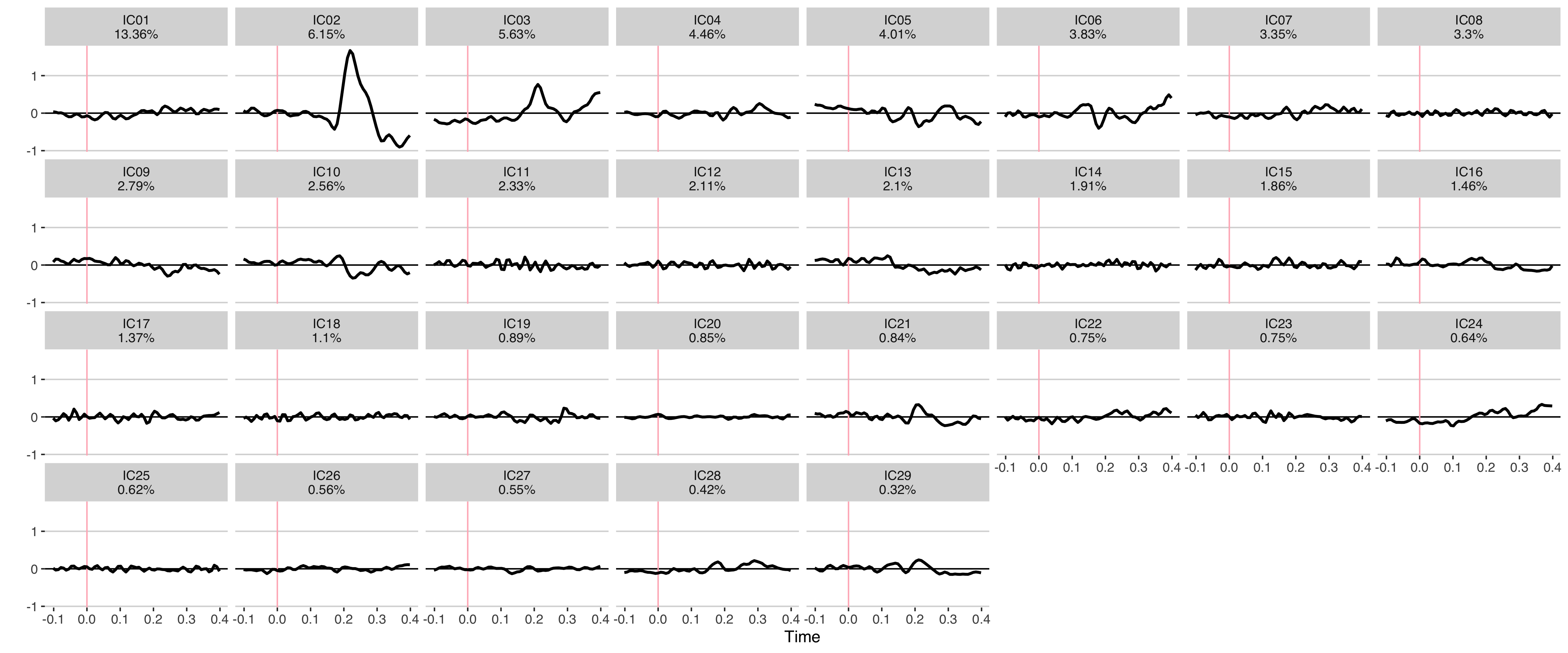}\\
\includegraphics[width=0.98\textwidth]{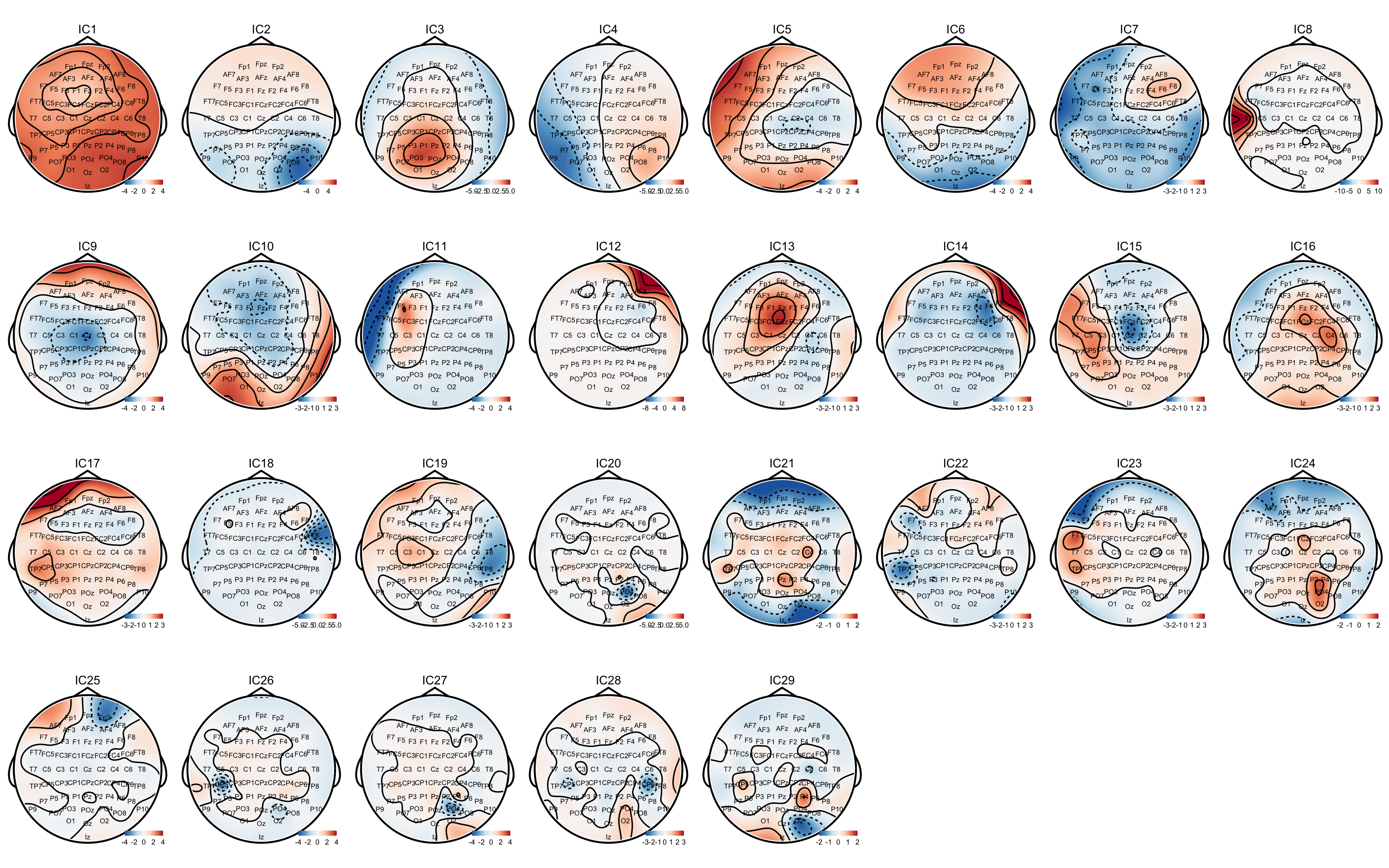}
\end{center}\caption{Topography maps of the 29 independent components identified by SOBI. ICs 1 and 5  were identified artifacts that are largely related to eye movement and mastoids.}
\label{fig:eegsobi}
\end{figure}

\begin{figure}\begin{center}
\includegraphics[width=\textwidth]{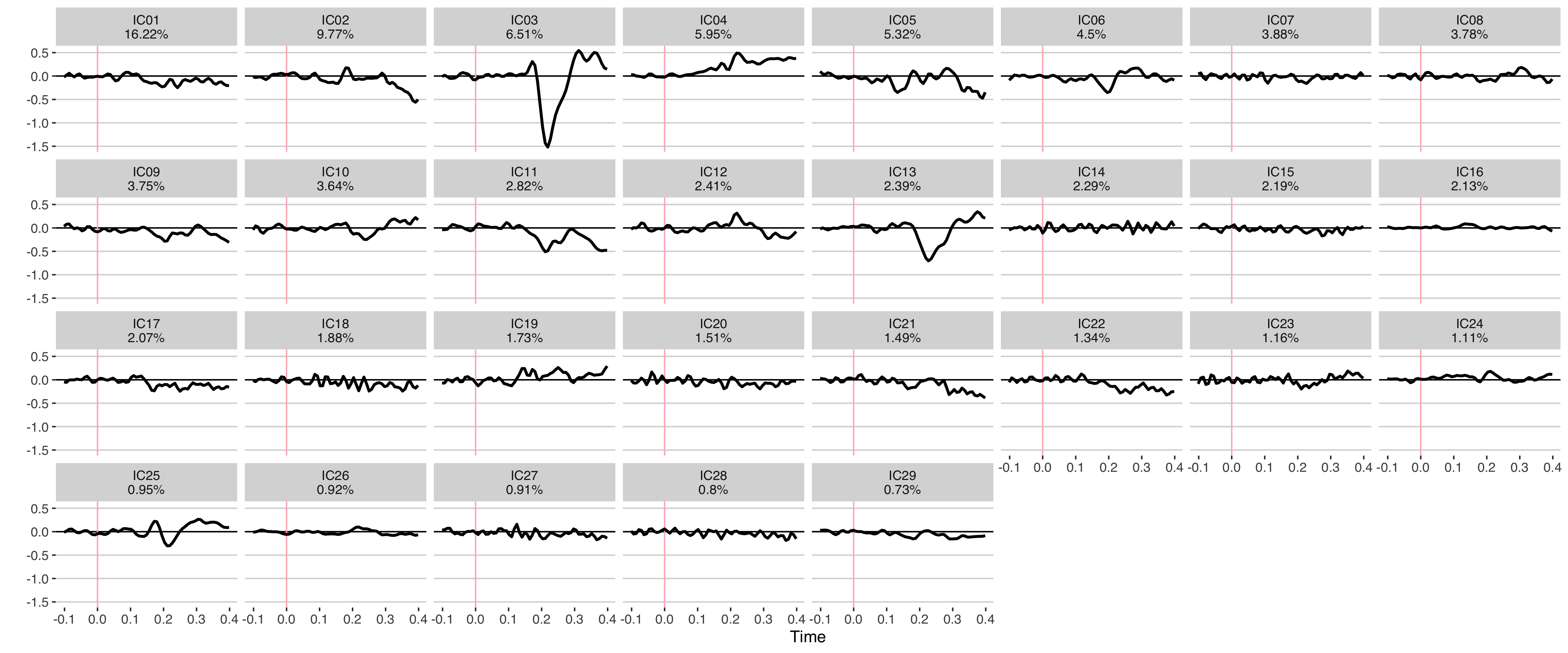}\\
\includegraphics[width=0.98\textwidth]{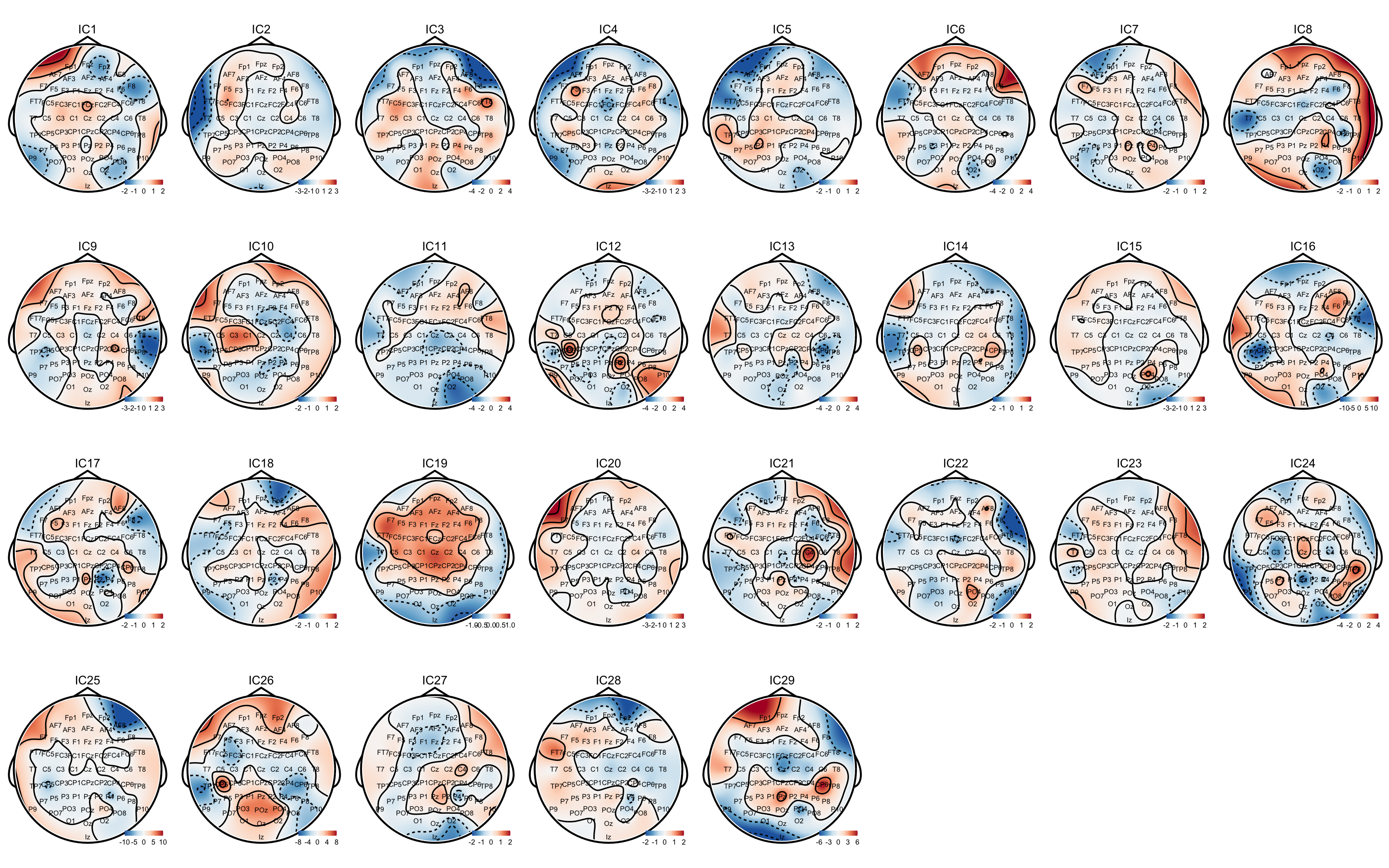}
\end{center}\caption{Topography maps of the 29 independent components identified by cICA-LSP. ICs 1, 4, 9 and 10  were identified artifacts that are largely related to eye movement.}
\label{fig:eeglsp}
\end{figure}

\begin{figure}\begin{center}
\includegraphics[width=0.9\textwidth]{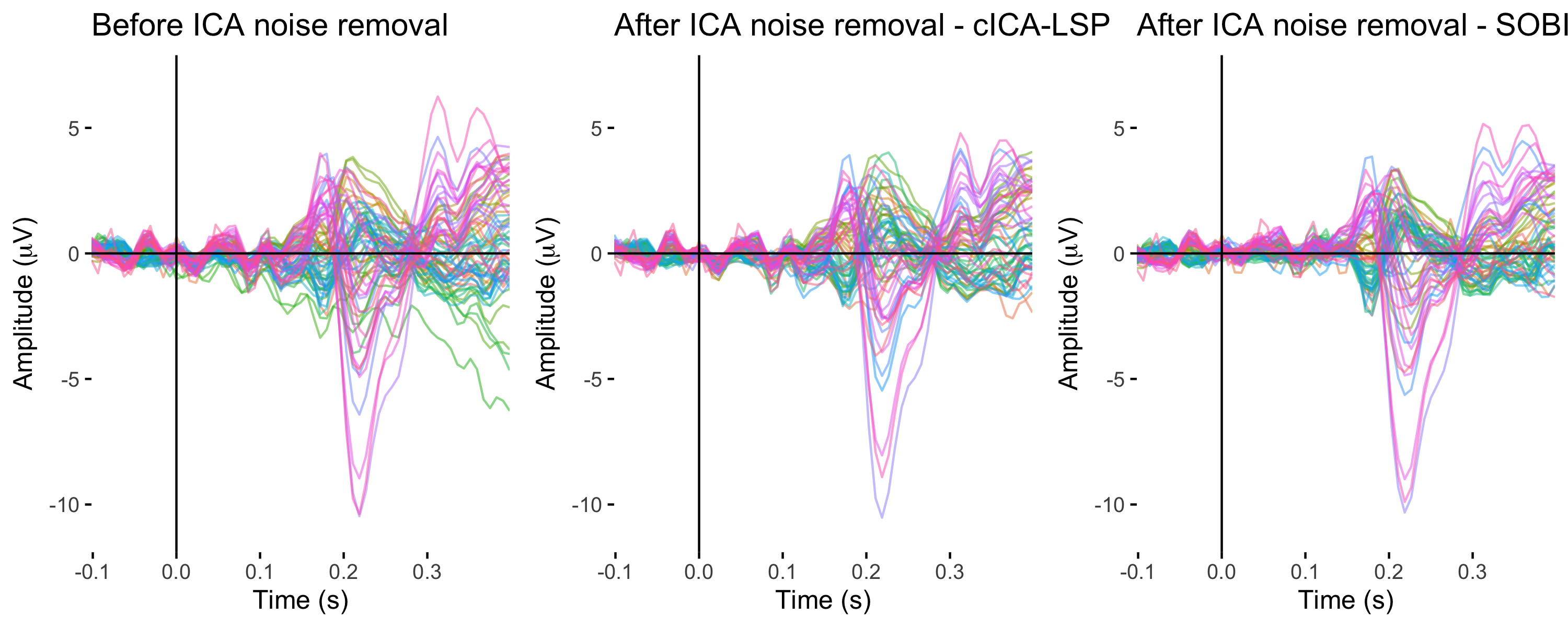}\\
\end{center}\caption{Average ERP before and after artifact removal at 64 electorodes using cICA-LSP and SOBI. }
\label{fig:eeglsp_removed}
\end{figure}

Here we included the rest of the EEG data analysis results. Figures \ref{fig:eegsobi} and \ref{fig:eeglsp} show the topography maps, average ERP, and variance explained by component of the 29 ICs obtained by SOBI and cICA-LSP. Figure \ref{fig:eeglsp_removed} shows the average ERP before and after artifact removal using SOBI (removing ICs 1 and 5) and cICA-LSP (removing ICs 1, 4, 9 and 10), respectively. 

\bibliography{references}

\begin{thebibliography}{47}
\providecommand{\natexlab}[1]{#1}
\providecommand{\url}[1]{\texttt{#1}}
\providecommand{\urlprefix}{}

\bibitem[{Hyv\"{a}rinen et~al.(2001)Hyv\"{a}rinen, A. and Karhunen, J and Oja,
  E.}]{hyvarinen2001ica}
Hyv\"{a}rinen A, Karhunen J, Oja E.
\newblock {Independent Component Analysis}.
\newblock John Wiley \& Sons; 2001.

\bibitem[{Hastie et~al.(2009)Hastie, T. and Tibshirani, R. and Friedman,
  J.H.}]{hastie2009elements}
Hastie T, Tibshirani R, Friedman JH.
\newblock {The Elements of Statistical Learning: Data Mining, Inference, and
  Prediction}.
\newblock Second ed. Springer Verlag; 2009.

\bibitem[{Cichocki et~al.(2009)Cichocki, A. and Zdunek, R. and Phan,
  A.H.}]{cichocki2009nonnegative}
Cichocki A, Zdunek R, Phan AH.
\newblock {Nonnegative Matrix and Tensor Factorizations: Applications to
  Exploratory Multi-way Data Analysis and Blind Source Separation}.
\newblock Wiley; 2009.

\bibitem[{Comon and Jutten(2010)Comon, P. and Jutten, C.}]{comon2010handbook}
Comon P, Jutten C.
\newblock {Handbook of Blind Source Separation: Independent Component Analysis
  and Applications}.
\newblock Academic Press; 2010.

\bibitem[{Back and Weigend(1997)Back, A.D. and Weigend, A.S.}]{back1997first}
Back AD, Weigend AS.
\newblock {A First Application of Independent Component Analysis to Extracting
  Structure from Stock Returns}.
\newblock International Journal of Neural Systems 1997;8:473--484.

\bibitem[{Calhoun et~al.(2009)Calhoun, V.D. and Liu, J. and Adali,
  T.}]{calhoun2009rgi}
Calhoun VD, Liu J, Adali T.
\newblock {A Review of Group ICA for fMRI Data and ICA for Joint Inference of
  Imaging, Genetic, and ERP Data}.
\newblock NeuroImage 2009;45(1S1):163--172.

\bibitem[{Makeig and Onton(2009)Makeig, S. and Onton, J.}]{makeig2009erp}
Makeig S, Onton J.
\newblock {ERP Features and EEG Dynamics: An ICA perspective}.
\newblock In: Luck S, Kappenman E, editors. {Oxford Handbook of Event-related
  Potential Components} Oxford University Press; 2009.

\bibitem[{Ge and Song(2007)Ge, Z. and Song, Z.}]{ge2007process}
Ge Z, Song Z.
\newblock {Process Monitoring Based on Independent Component Analysis-principal
  Component Analysis (ICA-PCA) and Similarity Factors}.
\newblock Industrial \& Engineering Chemistry Research 2007;46(7):2054--2063.

\bibitem[{Cardoso and Souloumiac(1993)Cardoso, J.F. and Souloumiac,
  A.}]{cardoso1993blind}
Cardoso JF, Souloumiac A.
\newblock Blind beamforming for non-Gaussian signals.
\newblock In: Radar and Signal Processing, IEE Proceedings F, vol. 140 IET;
  1993. p. 362--370.

\bibitem[{Bell and Sejnowski(1995)Bell, A.J. and Sejnowski, T.J.}]{bell1995ima}
Bell AJ, Sejnowski TJ.
\newblock {An Information-Maximization Approach to Blind Separation and Blind
  Deconvolution}.
\newblock Neural Computation 1995;7(6):1129--1159.

\bibitem[{Lee et~al.(1999)Lee, T.W. and Girolami, M. and Sejnowski,
  T.J.}]{lee1999independent}
Lee TW, Girolami M, Sejnowski TJ.
\newblock {Independent Component Analysis Using an Extended Infomax Algorithm
  for Mixed Subgaussian and Supergaussian Sources}.
\newblock Neural Computation 1999;11(2):417--441.

\bibitem[{Vlassis and Motomura(2001)Vlassis, N. and Motomura,
  Y.}]{vlassis2001esa}
Vlassis N, Motomura Y.
\newblock {Efficient Source Adaptivity in Independent Component Analysis}.
\newblock IEEE Transactions on Neural Networks 2001;12(3):559--566.

\bibitem[{Chen and Bickel(2006)Chen, A. and Bickel, P.J.}]{chen2006eic}
Chen A, Bickel PJ.
\newblock {Efficient Independent Component Analysis}.
\newblock The Annals of Statistics 2006;34(6):2825--2855.

\bibitem[{Bach and Jordan(2003)Bach, F.R. and Jordan, M.I.}]{bach2003kic}
Bach FR, Jordan MI.
\newblock {Kernel Independent Component Analysis}.
\newblock Journal of Machine Learning Research 2003;3(1):1--48.

\bibitem[{Boscolo et~al.(2004)Boscolo, R. and Pan, H. and Roychowdhury,
  VP}]{boscolo2004ica}
Boscolo R, Pan H, Roychowdhury V.
\newblock {Independent Component Analysis Based on Nonparametric Density
  Estimation}.
\newblock IEEE Transactions on Neural Networks 2004;15(1):55--65.

\bibitem[{Chen(2006)Chen, A.}]{chen2006fkd}
Chen A.
\newblock {Fast Kernel Density Independent Component Analysis}.
\newblock Independent Component Analysis and Blind Signal Separation
  2006;3889:24--31.

\bibitem[{Hastie and Tibshirani(2003)Hastie, T. and Tibshirani,
  R.}]{hastie2003ica}
Hastie T, Tibshirani R.
\newblock {Independent Components Analysis Through Product Density Estimation}.
\newblock In: Becker S, Obermayer K, editors. Advances in Neural Information
  Processing Systems, vol.~15 Cambridge, MA: MIT Press; 2003. p. 665--672.

\bibitem[{Kawaguchi and Truong(2009)Kawaguchi, A. and Truong, Y.
  K.}]{kawaguchi2007sic}
Kawaguchi A, Truong YK.
\newblock {Spline Independent Component Analysis}.
\newblock University of North Carolina at Chapel Hill; 2009.

\bibitem[{Eriksson and Koivunen(2003)Eriksson, J. and Koivunen,
  V.}]{eriksson2003cfb}
Eriksson J, Koivunen V.
\newblock {Characteristic-function-based Independent Component Analysis}.
\newblock Signal Processing 2003;83(10):2195--2208.

\bibitem[{Chen and Bickel(2005)Chen, A. and Bickel, PJ}]{chen2005cic}
Chen A, Bickel P.
\newblock {Consistent Independent Component Analysis and Prewhitening}.
\newblock IEEE Transactions on Signal Processing 2005;53(10 Part 1):3625--3632.

\bibitem[{Matteson and Tsay(2011)Matteson, D.S. and Tsay,
  R.S.}]{matteson2011independent}
Matteson DS, Tsay RS.
\newblock Independent Component Analysis via Distance Covariance.
\newblock Cornell University; 2011.

\bibitem[{Matteson and Tsay(2017)Matteson, David S and Tsay, Ruey
  S}]{matteson2017independent}
Matteson DS, Tsay RS.
\newblock Independent component analysis via distance covariance.
\newblock Journal of the American Statistical Association
  2017;112(518):623--637.

\bibitem[{Belouchrani et~al.(1997)Belouchrani, Adel and Abed-Meraim, Karim and
  Cardoso, J-F and Moulines, Eric}]{belouchrani1997blind}
Belouchrani A, Abed-Meraim K, Cardoso JF, Moulines E.
\newblock A blind source separation technique using second-order statistics.
\newblock IEEE Transactions on signal processing 1997;45(2):434--444.

\bibitem[{Miettinen et~al.(2014)Miettinen, Jari and Nordhausen, Klaus and Oja,
  Hannu and Taskinen, Sara}]{miettinen2014deflation}
Miettinen J, Nordhausen K, Oja H, Taskinen S.
\newblock Deflation-based separation of uncorrelated stationary time series.
\newblock Journal of Multivariate Analysis 2014;123:214--227.

\bibitem[{Pham and Garat(1997)Pham, D.T. and Garat, P.}]{pham1997bso}
Pham DT, Garat P.
\newblock {Blind Separation of Mixture of Independent Sources through a
  Quasi-maximum Likelihood Approach}.
\newblock IEEE Transactions on Signal Processing 1997;45(7):1712--1725.

\bibitem[{Lee et~al.(2011)Lee, S. and Shen, H. and Truong, Y.K. and Lewis, M.M.
  and Huang, X.}]{lee2011ica}
Lee S, Shen H, Truong YK, Lewis MM, Huang X.
\newblock {Independent Component Analysis Involving Autocorrelated Sources with
  an Application to functional Magnetic Resonance Imaging}.
\newblock Journal of the American Statistical Association 2011;106:1009--1024.

\bibitem[{Lee et~al.(2020)Lee, Seonjoo and Shen, Haipeng and Truong,
  Young}]{lee2020sampling}
Lee S, Shen H, Truong Y.
\newblock Sampling properties of color Independent Component Analysis.
\newblock Journal of Multivariate Analysis 2020;181:104692.

\bibitem[{Kooperberg et~al.(1995)Kooperberg, C. and Stone, C.J. and Truong,
  Y.K.}]{kooperberg1995lep}
Kooperberg C, Stone CJ, Truong YK.
\newblock {Logspline Estimation of a Possibly Mixed Spectral Distribution}.
\newblock Journal of Time Series Analysis 1995;16(4):359--388.

\bibitem[{Dzhaparidze and Kotz(1986)Dzhaparidze, K. and Kotz,
  S.}]{dzhaparidze1986peh}
Dzhaparidze K, Kotz S.
\newblock {Parameter Estimation and Hypothesis Testing in Spectral Analysis of
  Stationary Time Series}.
\newblock Springer; 1986.

\bibitem[{Brillinger(2001)Brillinger, D.R.}]{brillinger2001tsd}
Brillinger DR.
\newblock {Time Series: Data Analysis and Theory}.
\newblock Second ed. Society for Industrial and Applied Mathematics; 2001.

\bibitem[{Wahba(1980)Wahba, G.}]{wahba1980asl}
Wahba G.
\newblock {Automatic Smoothing of the Log Periodogram}.
\newblock Journal of the American Statistical Association
  1980;75(369):122--132.

\bibitem[{Chow and Grenander(1985)Chow, Y.S. and Grenander, U.}]{chow1985sms}
Chow YS, Grenander U.
\newblock {A Sieve Method for the Spectral Density}.
\newblock The Annals of Statistics 1985;13(3):998--1010.

\bibitem[{Franke and H{\"a}rdle(1992)Franke, J. and H{\"a}rdle,
  W.}]{franke1992bootstrapping}
Franke J, H{\"a}rdle W.
\newblock {On Bootstrapping Kernel Spectral Estimates}.
\newblock The Annals of Statistics 1992;20(1):121--145.

\bibitem[{Pawitan and O'Sullivan(1994)Pawitan, Y. and O'Sullivan,
  F.}]{pawitan1994nsd}
Pawitan Y, O'Sullivan F.
\newblock {Nonparametric Spectral Density Estimation Using Penalized Whittle
  Likelihood}.
\newblock Journal of the American Statistical Association
  1994;89(426):600--610.

\bibitem[{Fan and Kreutzberger(2001)Fan, J. and Kreutzberger,
  E.}]{fan1998automatic}
Fan J, Kreutzberger E.
\newblock {Automatic Local Smoothing for Spectral Density Estimation}.
\newblock Scandinavian Journal of Statistics 2001;25(2):359--369.

\bibitem[{Edelman et~al.(1998)Edelman, A. and Arias, T.A. and Smith,
  S.T.}]{edelman1998geometry}
Edelman A, Arias TA, Smith ST.
\newblock {The Geometry of Algorithms with Orthogonality Constraints}.
\newblock SIAM Journal on Matrix Analysis and Applications 1998;20:303--353.

\bibitem[{Amari(1999)Amari, S.I.}]{amari1999natural}
Amari SI.
\newblock {Natural Gradient Learning for Over-and Under-complete Bases in ICA}.
\newblock Neural Computation 1999;11(8):1875--1883.

\bibitem[{Douglas(2002)Douglas, S.C.}]{douglas2002self}
Douglas SC.
\newblock {Self-stabilized Gradient Algorithms for Blind Source Separation with
  Orthogonality Constraints}.
\newblock IEEE Transactions on Neural Networks 2002;11(6):1490--1497.

\bibitem[{Plumbley(2004)Mark D. Plumbley}]{plumbley2004lie}
Plumbley MD.
\newblock {Lie Group Methods for Optimization with Orthogonality Constraints}.
\newblock Independent Component Analysis and Blind Signal Separation
  2004;3195:1245--1252.

\bibitem[{Ye et~al.(2006)Ye, M. and Fan, X. and Liu, Q.}]{ye2006monotonic}
Ye M, Fan X, Liu Q.
\newblock {Monotonic Convergence of a Nonnegative ICA Algorithm on Stiefel
  Manifold}.
\newblock Lecture Notes in Computer Science 2006;4232:1098--1106.

\bibitem[{Schwarz(1978)Schwarz, G.}]{schwarz1978edm}
Schwarz G.
\newblock {Estimating the Dimension of a Model}.
\newblock The Annals of Statistics 1978;6(2):461--464.

\bibitem[{Amari et~al.(1996)Amari, S. and Cichocki, A. and Yang,
  H.H.}]{amari1996nla}
Amari S, Cichocki A, Yang HH.
\newblock {A New Learning Algorithm For Blind Signal Separation}.
\newblock Advances in Neural Information Processing Systems 1996;8:757--763.

\bibitem[{Chang(1999)Chang, J.}]{chang1999composing}
Chang J.
\newblock Composing Noise.
\newblock Institute of Sonology; 1999.

\bibitem[{Dinno(2009)Dinno, Alexis}]{dinno2009exploring}
Dinno A.
\newblock Exploring the sensitivity of Horn's parallel analysis to the
  distributional form of random data.
\newblock Multivariate behavioral research 2009;44(3):362--388.

\bibitem[{Ilmonen and Paindaveine(2011)Ilmonen, Pauliina and Paindaveine,
  Davy}]{ilmonen2011semiparametrically}
Ilmonen P, Paindaveine D.
\newblock Semiparametrically Efficient Inference Based on Signed Ranks in
  Symmetric Independent Component Models.
\newblock The Annals of Statistics 2011;39(5):2448--2476.

\bibitem[{Samworth and Yuan(2012)Samworth, Richard J and Yuan,
  Ming}]{samworth2012independent}
Samworth RJ, Yuan M.
\newblock Independent Component Analysis via Nonparametric Maximum Likelihood
  Estimation.
\newblock The Annals of Statistics 2012;40(6):2973--3002.

\bibitem[{Kooperberg et~al.(1995)Kooperberg, C. and Stone, C.J. and Truong,
  Y.K.}]{kooperberg1995rcl}
Kooperberg C, Stone CJ, Truong YK.
\newblock {Rate of Convergence for Logspline Sepctral Density Estimation}.
\newblock Journal of Time Series Analysis 1995;16(4):389--401.

\end{thebibliography}
\end{document}